\documentclass{article}
\usepackage{graphicx} % for inserting images
\usepackage{amsmath}
\usepackage{amssymb}
\usepackage{tikz}
\usetikzlibrary{shapes.geometric, arrows.meta, positioning, calc, fit}
\usepackage[hidelinks]{hyperref}
\usepackage{float}

\usepackage{tocloft}

\usepackage{fancyhdr}
\title{Knowledge, Rules and Their Embeddings: \\ Two\thanks{While some research may prefer a single-channel, universal or unified path, we believe in, and encourage, two and possibly more - any variants are welcome, if you find this work interesting.} Paths towards Neuro-Symbolic JEPA}
\author{Yongchao Huang\footnote{Email: yongchao.huang@abdn.ac.uk}, \hspace{0.1cm} Hassan Raza}
\date{20/02/2026}

\begin{document}

\maketitle

\begin{abstract}
Modern self-supervised predictive architectures excel at capturing complex statistical correlations from high-dimensional data but lack mechanisms to internalize verifiable human logic, leaving them susceptible to spurious correlations and shortcut learning. Conversely, traditional rule-based inference systems offer rigorous, interpretable logic but suffer from discrete boundaries and NP-hard combinatorial explosion. To bridge this divide, we propose a bidirectional neuro-symbolic framework centered around Rule-informed Joint-Embedding Predictive Architectures (RiJEPA). In the first direction, we inject structured inductive biases into JEPA training via Energy-Based Constraints (EBC) and a multi-modal dual-encoder architecture. This fundamentally reshapes the representation manifold, replacing arbitrary statistical correlations with geometrically sound logical basins. In the second direction, we demonstrate that by relaxing rigid, discrete symbolic rules into a continuous, differentiable logic, we can bypass traditional combinatorial search for new rule generation. By leveraging gradient-guided Langevin diffusion within the rule energy landscape, we introduce novel paradigms for continuous rule discovery, which enable unconditional joint generation, conditional forward and abductive inference, and marginal predictive translation. Empirical evaluations on both synthetic topological simulations and a high-stakes clinical use case confirm the efficacy of our approach. While classic JEPAs learn continuous features that require ex-post supervised downstream classifiers, RiJEPA intrinsically clusters its manifold during pretraining, achieving 100\% zero-shot logical accuracy and providing a transparent, distance-based geometric justification for its predictions. Ultimately, this framework establishes a powerful foundation for robust, generative, and interpretable neuro-symbolic representation learning.
\end{abstract}

\tableofcontents
\newpage

\section{Introduction} \label{sec:intro}

Modern deep learning, e.g. self-supervised predictive architectures, excels at capturing complex statistical correlations from high-dimensional, noisy data. However, these models act largely as black boxes; they lack inherent mechanisms to explicitly internalize structured human knowledge or adhere to verifiable logic. Conversely, traditional rule-based inference systems, such as Association Rule Mining (ARM) \cite{zhang2002association} and fuzzy logic system \cite{zadeh1965fuzzy}, offer rigorous, interpretable symbolic logic but suffer from rigid discrete boundaries, high sensitivity to arbitrary thresholds, and NP-hard combinatorial explosions when scaling to real-world complexity \cite{Zheng2001ARM}. The optimal paradigm may lie in a synthesis of these two extremes: merging the robust, continuous representation power of neural networks with the rigorous, compositional nature of symbolic logic.

Joint-Embedding Predictive Architectures (JEPAs) \cite{lecun2022path} have emerged as a powerful foundation for latent-space representation learning, filtering out low-level stochasticity to focus on high-level semantic features. Yet, current JEPAs remain purely data-driven. They map observations to representations based solely on statistical proximity, leaving them vulnerable to spurious correlations, shortcut learning, and out-of-distribution failures.

In this work, we address these limitations by investigating the \textit{bidirectional} synthesis of continuous predictive manifolds and discrete symbolic logic. Specifically, we aim to answer two core questions: 
\begin{enumerate}
    \item \textit{Direction 1 (Rules for JEPA):} can rule-informed training inject explicit structural inductive biases into JEPAs to geometrically shape the latent space, thereby anchoring representations to logical truths and improving out-of-distribution (OOD) generalization?
    \item \textit{Direction 2 (JEPA for Rules):} does converting discrete, NP-hard rule spaces into continuous, differentiable embedding manifolds enable new gradient-based paradigms for generative rule discovery, bypassing traditional combinatorial bottlenecks?
\end{enumerate}

To answer these questions, we introduce a unified neuro-symbolic framework that fundamentally transforms how predictive architectures interact with symbolic knowledge. By mapping foundational rule structures into the latent space via energy constraints, we create a landscape where logical validity equates to geometric proximity, enabling complex, zero-shot abductive reasoning.
To this end, this work makes four primary contributions to the fields of representation learning and neuro-symbolic AI:
\begin{itemize}
    \item \textit{Rule-Based JEPA (RbJEPA):} a novel methodology for distilling pure symbolic knowledge into continuous latent spaces.
    \item \textit{Rule-informed JEPA (RiJEPA) and Energy-Based Constraints (EBC):} a hybrid training objective that injects explicit structural inductive biases into raw-data training, actively mitigating shortcut learning by carving logically sound energy basins in the representation manifold.
    \item \textit{Multi-Modal Dual-Encoder Architecture:} a framework specifically designed to bridge the dimensional and semantic gap between high-dimensional continuous observations and discrete, low-dimensional symbolic logic.
    \item \textit{Continuous Rule Discovery Paradigms:} the introduction of two novel, generative rule discovery paradigms utilizing Langevin diffusion and the JEPA predictive module, replacing NP-hard combinatorial search with efficient, gradient-guided manifold exploration ($\nabla_z E(z)$).
\end{itemize}

\section{Related Work} \label{sec:related_work}

This work sits at the intersection of several research fields: self-supervised predictive architectures, neuro-symbolic AI, metric representation learning, and association rule mining. Here, we non-exhaustively review the foundational literature across these domains.

\subsection{Joint Embedding Predictive Architectures (JEPA)}
Joint-Embedding Predictive Architectures (JEPAs) have recently emerged as a powerful paradigm for self-supervised representation learning \cite{lecun2022path}. Unlike traditional generative models or autoencoders that rely on pixel-level or token-level reconstruction, JEPAs predict the latent representations of masked target regions from encoded context regions. This \textit{latent-to-latent} approach inherently filters out low-level stochasticity, focusing the alignment of high-level semantic features. 

Since the conceptual introduction of Hierarchical JEPA (H-JEPA) \cite{lecun2022path}, numerous variants have proven highly effective. I-JEPA \cite{assran2023ijepa} demonstrated highly semantic visual feature extraction without data augmentations, utilizing an asymmetric Exponential Moving Average (EMA) \cite{mnih2015human,mnih2016asynchronousmethodsdeepreinforcement,lillicrap2019continuouscontroldeepreinforcement,grill2020byol} encoder to prevent representation collapse. V-JEPA \cite{bardes2024vjepa} extended this to the temporal domain, learning motion dynamics via strictly causal predictions. Building upon these, JEPA World Models (JEPA-WMs) \cite{terver2025jepaworldmodels} and Value-Guided JEPA \cite{Destrade2026ValueGuidedJEPA} formalized the architecture for latent-space planning, optimizing action sequences directly within the embedding space without decoding to observations. 

While recent advances have introduced probabilistic formulations (e.g., VJEPA \cite{huang2026vjepavariationaljointembedding}) and bidirectional constraints (e.g., BiJEPA \cite{huang2026bijepa}), current JEPAs remain fundamentally data-driven. They rely on the statistical correlation of continuous inputs and lack mechanisms to explicitly internalize human-elicited logic or discrete structural priors.

\subsection{Neuro-Symbolic AI}
Neuro-symbolic AI aims to bridge the gap between connectionist learning (which excels at robust pattern recognition in noisy data) \cite{Hinton1990Connectionist} and symbolic logic (which offers rigorous, interpretable, and compositional reasoning) \cite{besold2017neuralsymbolic, davila2015neuralsymbolic}. Historically, early neuro-symbolic systems focused on knowledge extraction from neural networks or injecting propositional logic into training via specialized loss functions \cite{davila2015neuralsymbolic,dash2022review}. 

More recently, differentiable logic frameworks have gained traction. For example, Evans and Grefenstette introduced Differentiable Inductive Logic Programming, which learns explanatory rules from noisy data by relaxing discrete logic semantics into continuous weights \cite{evans2018learning}. Other contemporary approaches include Logic Tensor Networks (LTNs) \cite{serafini2016logic,Serafini2017LTN,Badreddine2022LTN}, which ground logical predicates in continuous spaces, and more recently hierarchical architectures that use Large Language Models as symbolic executives alongside continuous sensory grounding agents \cite{grosvenor2025hierarchical}. In the context of complex reasoning, neuro-symbolic architectures have also been adapted for knowledge-intensive question answering by executing logical compositions over parsed abstract structures \cite{cao2021general}. Further, within the domain of structured relational data, neural-symbolic integration has fundamentally advanced Knowledge Graph (KG) reasoning \cite{cheng2024neuralsymbolic}: contemporary KG methods successfully merge the robust representation power of neural embeddings with the compositional interpretability of symbolic rules. These approaches span from using logical rules as structural regularizers for continuous graph embeddings to developing fully differentiable rule-learning frameworks that execute logical queries end-to-end.

Our framework aligns with the neuro-symbolic goal of differentiable logic but shifts the paradigm from purely supervised or discriminative models to self-supervised joint-embedding spaces. Instead of using rules merely to regularize a final classifier, our proposed \textit{RiJEPA} utilizes them as explicit geometric constraints to shape the foundational representation manifold itself.

\subsection{Metric Learning and Latent Space Shaping}
To strictly enforce logical boundaries within continuous manifolds, our architecture heavily leverages concepts from deep metric learning \cite{Jain2008deepmetricslearning,Hoffer2015,wang2019multisimilaritylossgeneralpair,mohan2023deepmetriclearningCV}. Metric learning\footnote{Metric learning aims to automatically learn task-specific distance functions from (weakly) supervised data, rather than relying on predefined metrics such as Euclidean or cosine distance, enabling improved performance in tasks such as classification, clustering, and retrieval, etc. Deep metric learning focuses on learning representation functions that map data samples into an embedding space where similarity can be effectively measured \cite{mohan2023deepmetriclearningCV}.} optimizes embedding spaces such that semantically similar items are drawn together while dissimilar items are repelled. 

A standard approach in this domain is the use of \textit{Anchor Loss} (or proxy anchor loss) \cite{Kim_2020_CVPR} or centroid-based alignment, where continuous embeddings are pulled towards fixed, predefined coordinate poles (anchors) representing specific classes or states. This establishes absolute semantic locations in the latent space. A more dynamic and relative approach is \textit{Contrastive Triplet Training}, popularized by Large Margin Nearest Neighbor (LMNN) \cite{weinberger2009distance} and networks such as FaceNet \cite{schroff2015facenet}. Triplet loss \cite{he2018tripletcenterlossmultiview3d,wang2019multisimilaritylossgeneralpair} evaluates the relative distance between three simultaneous items: an Anchor ($A$), a Positive sample ($P$), and a Negative sample ($N$). It enforces the constraint that $d(A, P) < d(A, N) - m$, where $m$ is a contrastive margin.

In our framework, the Energy-Based Constraint ($\mathcal{L}_{EBC}$) operates on similar contrastive principles. By minimizing the energy (distance) of valid rule pairings and actively maximizing the energy of corrupted negative rule combinations via a margin, we geometrically sculpt the JEPA latent space into distinct, logically sound basins of attraction.

\subsection{Association Rule Mining and Rule Discovery}
The traditional method for discovering discrete symbolic logic from raw transactional data is Association Rule Mining (ARM) \cite{Agrawal1993}. Foundational algorithms such as \textit{Apriori} \cite{agrawal1994fast}, \textit{FP-Growth} \cite{han2000mining}, and \textit{Eclat} \cite{Zaki1997Eclat} utilize frequency-based measures (support, confidence, and lift) to navigate the dataset. Apriori employs a breadth-first search with candidate generation, Eclat utilizes depth-first search on vertical data layouts, and FP-Growth bypasses candidate generation entirely via compressed prefix trees. 

Despite their widespread use across retail, healthcare, and finance, traditional ARM approaches face critical limitations: they are highly sensitive to support thresholds, struggle to capture rare but highly significant associations, and generate massive volumes of redundant rules \cite{Pei2009}. Empirical evaluations on real-world datasets have starkly illustrated these bottlenecks; notably, Zheng and Kohavi \cite{Zheng2001ARM} demonstrated that even minor adjustments to minimum support thresholds can trigger a super-exponential explosion in rule generation. Such explosions rapidly produce millions of rules - far exceeding practical human utility and rendering raw algorithmic speed improvements largely irrelevant in applied settings. These limitations have spurred the development of Probabilistic and Bayesian ARM frameworks, such as Bayesian Association Rule (BAR) mining \cite{Tian2013BAR} and Bayesian Rule Mining (BRM) \cite{Gonzalez2020BRM}, which incorporate prior knowledge and evaluate belief rather than strict frequency. More recently, reinforcement learning and bandit-based frameworks (e.g., GIM-RL \cite{GIM2022} and MAB-ARM \cite{huang2025PARM}) have been introduced to dynamically adapt search strategies and optimize for high-quality, diverse rule sets.

\textbf{Our motivation.} While Bayesian and RL-based ARM methods introduce flexibility, they still operate within the rigid, NP-hard constraints of discrete combinatorial search spaces. Conversely, modern predictive architectures such as JEPA learn rich, continuous latent manifolds but remain blind to explicit symbolic constraints (i.e. lack of logic verification), and often rely purely on noisy statistical correlations. Our work provides a bidirectional framework that systematically resolves both gaps. To address the limitations of purely data-driven JEPAs (Direction 1), our \textit{Multi-Modal Dual-Encoder} and \textit{RiJEPA} framework utilize\footnote{It should be noted, however, that extracting optimal (verifiable and rational) rules from data or devising new ARM methods is out of the scope of this research; interested readers are referred to e.g., \cite{huang2025PARM}. In the following sections, we assume that logical rules are either readily extractable from the underlying raw dataset or provided as external prior knowledge (e.g., by human experts), such that they are ready to serve as a supervisory signal.} traditional ARM (e.g., FP-Growth) to establish foundational structural priors. By injecting these rules via Energy-Based Constraints ($\mathcal{L}_{EBC}$), we explicitly sculpt the JEPA latent space, imposing a structural inductive bias that mitigates shortcut learning and grounds the continuous manifold in verifiable, human-interpretable logic. Conversely, to address the combinatorial bottlenecks of traditional ARM (Direction 2), this geometrically shaped manifold transforms rigid discrete symbols into a differentiable energy landscape. This allows us to entirely bypass combinatorial search, replacing it with continuous, gradient-guided manifold exploration ($\nabla_z E(z)$). To our knowledge, no existing framework unifies self-supervised JEPA representation learning with discrete rule logic to enable such bidirectional, generative, and zero-shot abductive reasoning.

\section{Rules for JEPA: RbJEPA and RiJEPA} \label{sec:rules_for_jepa}

In this section, we explore how rules can be utilized to train, improve, or shape predictive architectures. We begin with a foundational concept, i.e. training a JEPA entirely on rules, before extending it to a hybrid formulation where rules act as auxiliary constraints for raw data.

\subsection{Rule-based JEPA (RbJEPA): Pure Symbolic Distillation} \label{subsec:RbJEPA}
RbJEPA represents the baseline scenario: a predictive model trained purely on logical rules extracted by an inference engine (e.g., decision trees, association rule mining, or fuzzy logic systems). The goal is to compress symbolic structure into continuous geometry.

\paragraph{Rule Extraction and Encoding}
Systematic extraction of rules from underlying datasets can be accomplished utilizing a variety of established techniques, e.g. decision trees, association rule mining (ARM), fuzzy inference systems (FIS), hierarchical clustering with pruning, and many other rule extraction frameworks. Regardless of the specific extraction method employed, each derived rule conforms to a standardized conditional structure:
\begin{center}
    \texttt{IF antecedent (conditions) $\rightarrow$ THEN consequent (outcome)}
\end{center}

Rather than processing flat text, each rule is converted into a structured object containing features, conditions, and statistical weightings (confidence, support). We construct meaningful embeddings by mapping the antecedent as the input to the context encoder $f_c$, and the consequent as the input to the target encoder $f_t$:
$$z_c = f_c(\text{antecedent}), \quad z_t = f_t(\text{consequent})$$

\paragraph{Training JEPA on Rule Sets}
We train the JEPA to predict the target representation directly from the context ($\hat{z}_t = g(z_c)$) by minimizing a weighted loss: 
\begin{equation}\label{eq:RbJEPA_loss}
  \mathcal{L}_{RbJEPA} = \sum_{i=1}^{N} w_i \|\hat z_t^{(i)} - z_t^{(i)}\|_2^2 
\end{equation}
where $w_i$ represents the statistical strength (e.g., confidence or support) of the extracted rule $i$. Unlike classical JEPA, which captures implicit statistical correlations from raw data observations, RbJEPA performs \textit{structured abstraction distillation}. It explicitly learns the underlying rule-level predictive structure of the domain, mapping discrete logical transitions into continuous vector transformations rather than merely memorizing target labels.

By adopting this formulation, we are effectively achieving three theoretical objectives: (i) compressing discrete symbolic knowledge into a continuous latent predictive space; (ii) grounding a foundational model in structured, causal-like abstractions rather than noisy input spaces; and (iii) utilizing verified logical rules as a robust, high-level mechanism for self-supervision.

Conceptually, this approach intersects with several existing paradigms, e.g. knowledge distillation \cite{hinton2015distilling, bucila2006model, gou2021knowledge}, neuro-symbolic learning \cite{davila2015neuralsymbolic,besold2017neuralsymbolic,mao2019neuro,evans2018learning}, relational contrastive learning \cite{zhang2023RCL, bordes2013translating, velickovic2019deep}, and structure-aware self-supervision \cite{lecun2022path, assran2023self, Lin2024structure}. However, leveraging the JEPA architecture introduces distinct advantages over traditional generative or autoregressive rule-learning models. First, because the architecture predicts latent embeddings rather than raw token outputs (e.g., text strings of rules), it naturally ignores low-level syntactic stochasticity. Second, rather than training the network to explicitly output an exact symbolic rule, we cultivate a generalized representation space. Within this space, similar antecedents cluster naturally, and valid rules operate as geometrically predictable transformations, this provides a more robust and powerful foundation for downstream reasoning tasks.

\paragraph{The Rule-Based JEPA Pipeline}
The proposed RbJEPA methodology is implemented as a three-step pipeline, as shown in Fig.~\ref{fig:rbjepa_pipeline}. Step 1 involves rule extraction. Step 2 encodes these rules into structured objects. Step 3 executes the JEPA training on these rule structures by minimizing a weighted predictive loss. 

\begin{figure}[H]
\centering
\begin{tikzpicture}[
    node distance=0.5cm,
    box/.style={rectangle, draw=black!70, rounded corners, fill=blue!2, thick, text width=0.85\textwidth, inner sep=8pt, align=justify, font=\small},
    arrow/.style={-{Stealth[length=3mm, width=2mm]}, thick, draw=black!70}
]

% Nodes
\node (step1) [box] {
    \textbf{Step 1: Rule Extraction} \\
    First, we mine a comprehensive rule set from the source data using any suitable rule-producing mechanism. Examples include a decision tree (e.g., \texttt{IF age > 50 AND cholesterol > 200 $\rightarrow$ risk = high}), association rules, or a fuzzy inference system. The output is a diverse collection of symbolic rules adhering to the form \texttt{IF antecedent $\rightarrow$ THEN consequent}.
};

\node (step2) [box, below=of step1] {
    \textbf{Step 2: Encoding Rules as Structured Objects} \\
    In the second step, each rule is systematically converted into a structured representation. For example:
    \vspace{-0.5em}
    {\footnotesize
    $$\text{Antecedent: } \{(feature_i, condition_i)\}, \quad \text{Consequent: } \{(output\_feature_j, value_j)\}$$
    $$\text{Confidence: } c,\quad \text{Support: } s.$$
    }
    \vspace{-1em} \\
    This formalization makes the extracted rules explicitly amenable to batching, embedding computation, and statistical weighting.
};

\node (step3) [box, below=of step2] {
    \textbf{Step 3: JEPA Training on Rule Structures} \\
    Finally, we treat the antecedent as the \emph{context} and the consequent as the \emph{target}. Let $f_c$ denote a context encoder, $f_t$ a target encoder, and $g$ a predictive module. For each rule $i$, we compute embeddings and predictions:
    \vspace{-0.5em}
    {\footnotesize
    $$z_c^{(i)} = f_c(\text{antecedent}^{(i)}),\qquad z_t^{(i)} = f_t(\text{consequent}^{(i)}),\qquad \hat z_t^{(i)} = g(z_c^{(i)})$$
    }
    \vspace{-1em} \\
    We minimize a weighted JEPA-style representation prediction loss summed over all rules (Eq.\ref{eq:RbJEPA_loss}):
    \vspace{-0.5em}
    {\footnotesize
    $$\mathcal{L}_{RbJEPA} = \sum_{i=1}^{N} w_i \|\hat z_t^{(i)} - z_t^{(i)}\|_2^2$$
    }
    \vspace{-1em} \\
    where $w_i$ denotes a statistical weighting factor derived from the empirical strength of rule $i$.
};

% Arrows
\draw [arrow] (step1) -- (step2);
\draw [arrow] (step2) -- (step3);

\end{tikzpicture}
\caption{The three-step pipeline for Rule-Based JEPA (RbJEPA), progressing from raw rule extraction to structured encoding and finally to representation learning via predictive loss minimization.}
\label{fig:rbjepa_pipeline}
\end{figure}

\subsection{Rule-informed JEPA (RiJEPA): Structured Inductive Bias} \label{subsec:RiJEPA}
While RbJEPA successfully distills pure logic, training a model purely on embedded rules has inherent limitations. Because rules are abstracted from raw data via a rule-based learning system, they typically summarize only the key information. Consequently, a purely rule-based model can easily miss or overlook nuanced yet critical relations present in the complex data. Conversely, conventional, purely data-driven JEPA often entangles noise and relies heavily on spurious correlations. 

To overcome the disadvantages of both extremes, we propose training a JEPA using multimodal sources. Specifically, we design a Rule-informed JEPA (RiJEPA) to inject a structured inductive bias. By presenting the JEPA simultaneously with raw data and extracted rules, this dual-channel approach combines raw observational signals with human-elicited knowledge or machine-extracted symbolic rules. This shifts the learning paradigm from unconstrained prediction to structurally grounded representation learning. The rules serve as an auxiliary modality, actively constraining and reshaping the JEPA's latent space to strictly conform to prior knowledge and known abstractions.

\paragraph{Energy-Based Constraints (EBC)}
Let $\mathcal{L}_{JEPA}$ denote the standard self-supervised prediction loss over raw data. RiJEPA augments this with an explicit rule-based constraint: 
\begin{equation} \label{eq:RiJEPA_loss}
    \mathcal{L}_{total} = \mathcal{L}_{JEPA} + \beta \mathcal{L}_{EBC}
\end{equation}
We implement this auxiliary objective via an Energy-Based formulation, defining the predicted energy of a symbolic rule as\footnote{This energy functional design is inspired by statistical physics and electrostatics, in which low-energy states are inherently more stable and attractive, while high-energy states are unstable and repulsive. By analogy, valid logical mappings are optimized to reside in stable, low-energy basins within the latent space, whereas invalid or contradictory rules are actively pushed into unstable, high-energy regions.} $E(A, C) = \|g(f_c(A)) - f_t(C)\|_2^2$.

By minimizing the predicted energy for valid rules and actively maximizing it for invalid pairs using a margin $m$, the model learns a structured rule validity landscape\footnote{A detailed theoretical comparison between this energy-based approach and the simple pairwise predictive supervision from Eq.~\ref{eq:RbJEPA_loss} is provided in Appendix \ref{app:PPS_vs_EBC}.}:
\begin{equation} \label{eq:RiJEPA_EBC_loss}
    \mathcal{L}_{EBC} = \sum_{\text{valid}} E(A, C) + \lambda \sum_{\text{invalid}} \max(0, m - E(A, C_{\text{neg}}))
\end{equation}
Here, $\beta$ balances the rule constraint against the raw data loss, while $\lambda$ balances the negative pair penalization against the positive pair minimization within the EBC itself. Intuitively, this objective acts as a geometric push-pull mechanism: it forces the predictive latent space to embed valid logical transitions as highly predictable, low-energy basins, while actively repelling contradictory pairs into high-energy regions until they are separated by at least the distance $m$. Invalid pairs $(A, C_{\text{neg}})$ may be generated dynamically during training by taking a valid antecedent $A$ and replacing its true consequent with randomly sampled or structure-preserving corrupted alternatives $C_{\text{neg}}$.

\paragraph{Supervising Relationships over Outputs}
Conceptually, RiJEPA fundamentally diverges from traditional knowledge distillation. While classical distillation focuses on mimicking a teacher model's final output labels, RiJEPA constrains the internal latent structure itself to align predictive geometry with symbolic abstractions. The system actively supervises the \emph{relationships} between entities rather than merely matching target outputs, thereby enforcing rigorous structural supervision over simple task supervision. 

\paragraph{Benefits for Alignment and Robustness}
From an alignment perspective, explicitly enforcing these energy constraints ensures that the model's latent space mathematically respects known domain boundaries. Consequently, this methodology actively mitigates the learning of spurious correlations and naturally biases the predictive geometry towards a more human-interpretable, causal structure. Ultimately, RiJEPA is designed to yield significant improvements in data efficiency, latent interpretability, reduction of shortcut learning, and OOD generalization.

\paragraph{RiJEPA: A Multi-Modal Dual-Encoder Architecture} \label{subsec:dual_encoder}
In simple environments, e.g. Section.\ref{subsec:simulated_example}, raw data and rules might share the same dimensionality (i.e. a single pair of context and target encoders ($f_c, f_t$) can process both data and rule modalities); however, in real-world neuro-symbolic systems, these modalities are fundamentally distinct:
\begin{itemize}
    \item \textit{Raw Data:} high-dimensional, continuous, and noisy (e.g., video frames of a car, multi-sensor states).
    \item \textit{Symbolic Rules:} low-dimensional, discrete, and abstract (e.g., logical graphs or text such as \texttt{IF red light $\rightarrow$ THEN stop}).
\end{itemize}

\begin{figure}[H]
    \centering
    \begin{tikzpicture}[
        node distance=0.5cm and 0.6cm, 
        encoder/.style={rectangle, draw=blue!80, fill=blue!10, rounded corners, text centered, minimum width=1.4cm, minimum height=0.7cm, font=\small},
        target_enc/.style={rectangle, draw=orange!80, fill=orange!10, rounded corners, text centered, minimum width=1.4cm, minimum height=0.7cm, font=\small},
        predictor/.style={rectangle, draw=green!60!black, fill=green!10, rounded corners, text centered, minimum width=1.6cm, minimum height=1.2cm, font=\small, thick},
        state/.style={circle, draw=black!70, fill=gray!10, minimum size=0.8cm, inner sep=0pt, font=\footnotesize},
        flow_line/.style={-{Stealth[length=2mm, width=1.5mm]}, thick, draw=black!70},
        stop_grad/.style={rectangle, draw=red!80, fill=red!10, font=\scriptsize, inner sep=1pt},
        label_text/.style={font=\small\bfseries, align=center, color=black!70}
    ]

    \node[predictor, align=center] (predictor_g) at (0, 0) {Shared\\Predictor $g(\cdot)$};

    \node[state, above left=0.4cm and 0.7cm of predictor_g] (z_xc) {$z_{c\_data}$};
    \node[encoder, left=0.6cm of z_xc] (enc_xc) {$f_{c\_data}$};
    \node[left=0.5cm of enc_xc, font=\small] (x_data) {Input $x$};

    \node[state, below left=0.4cm and 0.7cm of predictor_g] (z_ac) {$z_{c\_rule}$};
    \node[encoder, left=0.6cm of z_ac] (enc_ac) {$f_{c\_rule}$};
    \node[left=0.5cm of enc_ac, font=\small] (a_rule) {Antecedent $A$};

    \draw[flow_line] (x_data) -- (enc_xc);
    \draw[flow_line] (enc_xc) -- (z_xc);
    \draw[flow_line] (a_rule) -- (enc_ac);
    \draw[flow_line] (enc_ac) -- (z_ac);
    
    \draw[flow_line, rounded corners] (z_xc.east) -- ++(0.2,0) |- ($(predictor_g.west)+(0,0.2)$);
    \draw[flow_line, rounded corners] (z_ac.east) -- ++(0.2,0) |- ($(predictor_g.west)-(0,0.2)$);

    \node[state, above right=0.4cm and 0.08cm of predictor_g] (z_yt_hat) {$\hat{z}_{t\_data}$};
    \node[state, right=0.7cm of z_yt_hat] (z_yt) {$z_{t\_data}$};
    \node[target_enc, right=1.2cm of z_yt] (enc_yt) {$f_{t\_data}$};
    \node[right=0.5cm of enc_yt, font=\small] (y_data) {Target $y$};

    \node[state, below right=0.4cm and 0.08cm of predictor_g] (z_ct_hat) {$\hat{z}_{t\_rule}$};
    \node[state, right=0.7cm of z_ct_hat] (z_ct) {$z_{t\_rule}$};
    \node[target_enc, right=1.2cm of z_ct] (enc_ct) {$f_{t\_rule}$};
    \node[right=0.5cm of enc_ct, font=\small] (c_rule) {Consequent $C$};

    \draw[flow_line] (y_data) -- (enc_yt);
    \draw[flow_line] (enc_yt) -- node[stop_grad] {SG} (z_yt);
    \draw[flow_line] (c_rule) -- (enc_ct);
    \draw[flow_line] (enc_ct) -- node[stop_grad] {SG} (z_ct);

    \draw[flow_line, rounded corners] ($(predictor_g.east)+(0,0.2)$) -- ++(0.2,0) -| (z_yt_hat.south);
    \draw[flow_line, rounded corners] ($(predictor_g.east)-(0,0.2)$) -- ++(0.2,0) -| (z_ct_hat.north);

    \node[label_text, above=0.1cm of enc_xc, xshift=-1.2cm] {Data Modality (Continuous)};
    \node[label_text, below=0.1cm of enc_ac, xshift=-1.2cm] {Rule Modality (Symbolic)};

    \draw[dashed, gray, thick, rounded corners] 
        ($(z_xc.north west)+(-0.25, 0.5)$) rectangle ($(z_ct.south east)+(0.25, -0.4)$);
    \node[font=\small\bfseries\color{gray}, fill=white, inner sep=3pt] at (0.5, 2.5) {Shared Latent Semantic Space};

    \draw[<->, dashed, red, thick] (z_yt_hat) -- node[midway, above, font=\footnotesize, text=red] {$\mathcal{L}_{JEPA}$} (z_yt);
    \draw[<->, dashed, red, thick] (z_ct_hat) -- node[midway, below, font=\footnotesize, text=red] {$\mathcal{L}_{EBC}$} (z_ct);

    \end{tikzpicture}
    \caption{The Multi-Modal Dual-Encoder Architecture bridging continuous data and discrete logic. Continuous data contexts ($x$) and symbolic rule antecedents ($A$) are encoded on the left and converge on a universal predictor $g$. The predictions are compared against target encodings of data ($y$) and consequents ($C$) generated on the right, enabling joint minimization of data loss ($\mathcal{L}_{JEPA}$) and rule energy constraints ($\mathcal{L}_{EBC}$). SG denotes \textit{stop gradient}.}
    \label{fig:dual_encoder}
\end{figure}
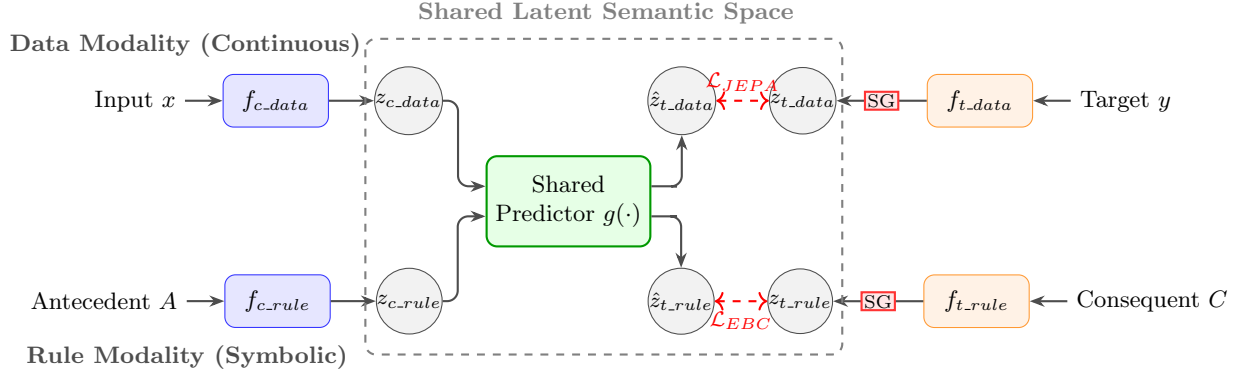

To generalize RiJEPA to complex environments where data and rule modalities don't generally share the same dimensions (and semantic scale), we propose a \textbf{Multi-Modal Dual-Encoder Architecture}. As shown in Fig.\ref{fig:dual_encoder}, this architecture utilizes four distinct encoders to project fundamentally different data types into a \textit{Shared Latent Semantic Space}:
\begin{enumerate}
    \item \textit{Data Encoders ($f_{c\_data}, f_{t\_data}$):} these are domain-specific architectures (e.g., CNNs or Vision Transformers) which map high-dimensional raw observations $(x, y)$ into latent representations $(z_{c\_data}, z_{t\_data})$. 
    
    \item \textit{Rule Encoders ($f_{c\_rule}, f_{t\_rule}$):} these are symbolic encoders (e.g., Text Transformers or Graph Neural Networks) that map discrete rule antecedents and consequents $(A, C)$ into latent representations $(z_{c\_rule}, z_{t\_rule})$.
    
    \item \textit{The Shared Universal Predictor ($g$):} a single predictive module operating strictly within the shared latent space. It is designed to be highly flexible: it can simultaneously accept both the raw data context embedding and the rule antecedent embedding to jointly predict the data target and rule consequent representations. If either modality is unavailable, for example, accepting only data embeddings or only rule embeddings during specific training phases or inference, the missing input is simply masked or zero-padded\footnote{This is similar to how masked autoencoders handle missing modalities.}. \textit{Notably, when rule embeddings are completely masked, the architecture gracefully reduces to a standard, purely data-driven JEPA}.
\end{enumerate}

Because the modalities share a unified latent space, the predictor $g$ acts as a universal bridge, enabling \textit{Zero-Shot Logical Inference}. By passing a raw data context through the predictor ($g(z_{c\_data})$), the output can be directly compared against a symbolic rule consequent space ($f_{t\_rule}(C)$) to answer queries such as: \textit{``Does the current observed state satisfy logical condition C?''}. For example, in our later experiment Section.\ref{subsec:clinical_case}, we establish fixed diagnostic ``poles'' by passing explicit clinical outcomes (e.g., \texttt{target\_risk=1.0} for high risk and \texttt{target\_risk=0.0} for low risk) through the frozen rule target encoder $f_{t\_rule}$. To diagnose a new, unseen patient, we project their raw physiological vitals through the data pathway to yield a predicted state $g(f_{c\_data}(x_{\text{test}}))$ and simply compute the Euclidean distance to these predefined symbolic poles. The patient is then classified based strictly on their spatial proximity to these logical coordinates, executing the prediction entirely zero-shot without the need for test-time rule mining or the training of a downstream classifier.

\vspace{1em}

% ==========================================
% JEPA FOR RULES
% ==========================================
\section{JEPA for Rules: Continuous Rule Manifolds Learning} \label{sec:jepa_for_rules}

While Section.\ref{sec:rules_for_jepa} demonstrated how symbolic rules act as priors for representations, this relationship is fundamentally bidirectional. The secondary goal of our framework is to utilize the resulting continuous, differentiable embedding space to execute gradient-based generative rule discovery, moving away from classical combinatorial search of new rules.

\subsection{From Combinatorial Search to Continuous Rule Manifolds}
Classical discrete rule discovery is effectively a ``combinatorial hell'' which is heavily reliant on non-differentiable, NP-hard search heuristics \footnote{Association rule mining \cite{huang2025PARM}, for example, heavily utilizes algorithms such as Apriori \cite{agrawal1994fast}, Eclat \cite{Zaki1997Eclat}, and FP-Growth \cite{han2000mining}; while decision tree induction relies on recursive partitioning methods like ID3 \cite{quinlan1986induction}, C4.5 \cite{quinlan1993c4}, and CART \cite{breiman1984cart}.}  (e.g., Apriori \cite{agrawal1994fast} or CART \cite{breiman1984cart}). By enforcing the Energy-Based Constraint (EBC) detailed previously, we fundamentally shift this paradigm. The discrete symbolic rule space is transformed into a smooth, continuous embedding space where rule validity is relaxed into an energy function: $E(z_c, z_t) = \|g(z_c) - z_t\|_2^2$.

The rule manifold itself can be formally defined as the sublevel set of this energy landscape:
$$\mathcal{M}_{rule} = \{(z_c, z_t) \mid E(z_c, z_t) \le \epsilon \}$$
In this framework, low-energy regions correspond to valid rules. Rules are no longer isolated, discrete symbolic objects; they are coordinate points residing in a differentiable manifold.

\subsection{Two Paradigms for Generative (New) Rule Discovery}
By relaxing logical boundaries into continuous landscapes, we can execute rule discovery using generative modeling techniques. Depending on the computational constraints, this can be executed via two distinct paradigms.

\paragraph{Paradigm 1: Joint Energy-Based Diffusion}
The first approach treats rule validity as a low-energy state defined by the joint probabilistic distribution\footnote{
As the energy landscape is defined as $E(z_c, z_t) = \|g(z_c) - z_t\|_2^2$, which is learned (via Eq.\ref{eq:RiJEPA_loss} and Eq.\ref{eq:RiJEPA_EBC_loss}) during training, here we are essentially turning a deterministic model into a probabilistic one, hypothesizing that it is more likely to observe data that are similar to those training data yielding smaller MSE. So rules generated via diffusion are essentially interpolations of the rules that are consistent with the training data.
}:
\begin{equation} \label{eq:joint_context_target_dist}
    p(z_c, z_t) = \frac{1}{Z} \exp\left(-\frac{E(z_c, z_t)}{T}\right)
\end{equation}
where $Z$ is a partition function and the energy is defined by the predictive compatibility $E(z_c, z_t) = \|g(z_c) - z_t\|_2^2$. We can execute continuous rule discovery by flowing through the joint rule space towards valid, low-energy configurations using gradient-guided \textit{Langevin dynamics}:
\begin{equation} \label{eq:Langevin_dynamics}
    z^{(k+1)} = z^{(k)} - \eta \nabla_z E(z^{(k)}) + \sqrt{2\eta T}\epsilon
\end{equation}
where $k$ denotes the diffusion step and $\epsilon \sim \mathcal{N}(0, I)$. In this formulation, $z$ represents the joint state of the rule, defined as the concatenated vector of both the antecedent and consequent embeddings: $z = [z_c, z_t]^\top$. Because the energy function $E(z_c, z_t)$ yields a scalar value over this joint space, taking the gradient $\nabla_z E(z)$ computes the partial derivatives with respect to both components simultaneously:
$$
\nabla_z E(z) = \begin{bmatrix} \nabla_{z_c} E(z_c, z_t) \\ \nabla_{z_t} E(z_c, z_t) \end{bmatrix}
$$

This joint continuous representation allows the diffusion process to be utilized in three distinct generative modes:
\begin{itemize}
    \item \textit{Unconditional Rule Discovery (Joint Diffusion):} Updating both $z_c$ and $z_t$ simultaneously using the full joint gradient, the system freely wanders down the energy landscape to hallucinate entirely new, logically valid rules from scratch according to:
    $$ \begin{bmatrix} z_c \\ z_t \end{bmatrix}_{k+1} = \begin{bmatrix} z_c \\ z_t \end{bmatrix}_k - \eta \nabla_{z_c, z_t} \|g(z_c) - z_t\|_2^2 + \sqrt{2\eta T}\epsilon $$
    
    \item \textit{Forward Inference (Conditional Diffusion):} Fixing the antecedent $z_c$ and exclusively applying Langevin dynamics to $z_t$, the model searches for the corresponding logical consequent via:
    $$ z_{t, k+1} = z_{t, k} - \eta \nabla_{z_t} \|g(z_c) - z_t\|_2^2 + \sqrt{2\eta T}\epsilon $$
    This effectively asks: \textit{``Given this condition, what is the valid outcome?''}
    
    \item \textit{Abductive Reasoning (Backward Inference):} Fixing the consequent $z_t$ and exclusively updating the context $z_c$, the model performs root-cause analysis via:
    $$ z_{c, k+1} = z_{c, k} - \eta \nabla_{z_c} \|g(z_c) - z_t\|_2^2 + \sqrt{2\eta T}\epsilon $$
    This effectively asks: \textit{``What antecedent conditions would lead to this specific outcome?''}
\end{itemize}

The discovery process is governed by the Langevin SDE where the update step is composed of two primary components. The \textit{Drift Term}, $-\eta \nabla \|g(z_c) - z_t\|_2^2$, represents the deterministic movement of the latent vectors towards the logical basins defined by the predictor $g$, which enforces the ``compatibility'' between the antecedent and consequent. Complementing this is the \textit{Diffusion Term}, $\sqrt{2\eta T}\epsilon$, which introduces Gaussian noise to allow the system to explore the manifold and escape non-logical local minima.

In this context, the temperature parameter $T$ is a critical hyperparameter that controls the ``creativity'' or entropy of the rule hallucinations. A low $T$ constrains the model to highly probable, conservative clinical associations, while a high $T$ increases the stochasticity, allowing the model to traverse wider regions of the latent space to discover more complex clinical patterns. By taking the gradient with respect to the Predictor $g$ in all three discovery tasks, we ensure that every generated rule is anchored in the learned structural logic of the domain.

\paragraph{Paradigm 2: Marginal-Predictive Guided Discovery}
While joint diffusion over $p(z_c, z_t)$ is mathematically elegant, modeling high-dimensional joint distributions can be computationally intensive and sensitive to landscape artifacts. To address this, we introduce an alternative paradigm that explicitly exploits the asymmetric, predictive nature of the JEPA architecture, decoupling generative exploration from deterministic prediction into three phases:
\begin{enumerate}
    \item \textit{Marginal Context Generation:} instead of learning the joint distribution, we model only the marginal distribution of the context embeddings (representing rule antecedents), denoted as $p(z_c) \propto \exp(-E_{ctx}(z_c))$. We use Langevin diffusion to generate novel, structurally valid contexts:
    $$z_c^{(\tau+1)} = z_c^{(\tau)} - \eta \nabla_{z_c} E_{ctx}(z_c^{(\tau)}) + \sqrt{2\eta}\epsilon$$
    
    \item \textit{Feed-Forward Target Prediction:} once a novel context embedding $z_c$ is generated, we exploit the deterministic strength of the JEPA architecture. We pass the generated context directly through the shared predictor to yield the implied consequent: $\hat{z}_t = g(z_c)$. Because $g$ has been strongly regularized by the EBC loss, this acts as a highly informed, zero-shot projection of the rule's outcome.
    
    \item \textit{Metropolis-Hastings (MH) Validation:}  to rigorously assess the quality of the newly proposed rule pair $(z_c, \hat{z}_t)$, the latent representations are decoded back into their symbolic forms $(\hat{A}, \hat{C})$. We then subject the proposed rule to an MH acceptance step against a target validity distribution, where it is accepted with probability:
    $$ \alpha = \min \left( 1, \frac{\pi(\hat{A}, \hat{C})}{\pi(A_{\text{old}}, C_{\text{old}})} \right) $$
    where $\pi(\cdot)$ is determined by an external knowledge base, a deterministic physics simulator, or explicit human-in-the-loop feedback to ensure structural logic is preserved.
\end{enumerate}
By restricting the diffusion process to the context marginal $p(z_c)$ and relying on $g(z_c)$ to compute the consequent, this paradigm significantly reduces the generative search space while strictly preserving the causal mapping learned during RiJEPA training.

\subsection{Differentiable Logic and Interpretability Requirements}

The energy minimization objective in RiJEPA (Eq.~\ref{eq:RiJEPA_EBC_loss}) enables scalable, continuous rule manifold learning. By embedding discrete symbolic components into continuous representations, we establish a \textit{differentiable logic} which allows the network to compute gradients through otherwise rigid rules to smoothly navigate and optimize the rule space. However, an important architectural subtlety must be met: the encoder cannot be an arbitrary multi-layer perceptron (MLP) operating on flat text. To safely map generative geometry back into understandable rules, the latent space must represent distinct semantic entities. 

To formalize this structural constraint, we define a comprehensive rule embedding tuple, $\mathcal{T}_{\text{rule}}$, which encapsulates the essential atomic components required for continuous manifold learning:
$$\mathcal{T}_{\text{rule}} = \langle X, \Phi, \Theta, Y, \mu, \Omega, \Sigma \rangle$$
where each element maps to a distinct geometric entity in the latent space:
\begin{itemize}
    \item $X$: Input features or raw data variables (e.g., \texttt{age}).
    \item $\Phi$: Relational conditions or predicates (e.g., $<, >, \leq$).
    \item $\Theta$: Numeric threshold values or specific categorical states.
    \item $Y$: Output features or target outcomes (e.g., \texttt{heart\_risk = high}).
    \item $\mu$: Side info such as membership degrees derived from fuzzy functions.
    \item $\Omega$: Logical operators (e.g., AND, OR, NOT).
    \item $\Sigma$: Statistical rule strength metrics (e.g., support, confidence, lift).
\end{itemize}

To illustrate, consider a medical diagnostic scenario where an extracted rule states: \texttt{IF age > 50 AND cholesterol > 200 $\rightarrow$ heart\_risk = high} with a confidence of 85\%. This rule explicitly populates the tuple as: $X = [\texttt{age}, \texttt{cholesterol}]$, $\Phi = [>, >]$, $\Theta = [50, 200]$, $\Omega = [\texttt{AND}]$, $Y = [\texttt{heart\_risk = high}]$, and $\Sigma = [0.85]$, with $\mu$ capturing any fuzzy membership degrees assigned by the rule engine.

A prominent risk of relaxing discrete logic into a generative energy landscape is the potential loss of interpretability. The network might converge on low-energy ``rules'' that exist as entangled, nonsensical mixtures in the latent space. Therefore, to safely utilize this generative logic, a robust latent-to-symbolic rule decoding projection mechanism is required:
$$\hat{\mathcal{T}}_{\text{rule}} = D(z)$$
This decoder $D(z)$ ensures interpretability by mapping latent representations back to their strict, decipherable symbolic structure. 

Further, enforcing this structured embedding natively unlocks advanced generative capabilities. First, it enables \textit{rule interpolation}, i.e. dynamically navigating the space to discover ``nearby'' valid rules (e.g., smoothly adjusting a numeric threshold $\Theta$ from \texttt{age > 50} to \texttt{age > 55}) that were never explicitly extracted from the training data. Second, \textit{symbolic compositionality} emerges directly from the representation geometry, allowing for composite rule discovery via continuous vector arithmetic (e.g., $z_{\text{Rule A}} + z_{\text{Rule B}} \approx z_{\text{new Rule C}}$).

\section{Experiments} \label{sec:experiments}

To demonstrate the geometric shaping properties of RiJEPA and new rule discovery, we present an analytical illustration, a computational simulation and a real-world example.

\subsection{Analytical Illustration: Simple Feature Rules}
Consider a straightforward system with two features $x_1, x_2$ and three valid symbolic rules:
$$R_1: x_1 > 0 \rightarrow y = 1$$
$$R_2: x_2 > 0 \rightarrow y = 1$$
$$R_3: x_1 \leq 0 \rightarrow y = 0$$

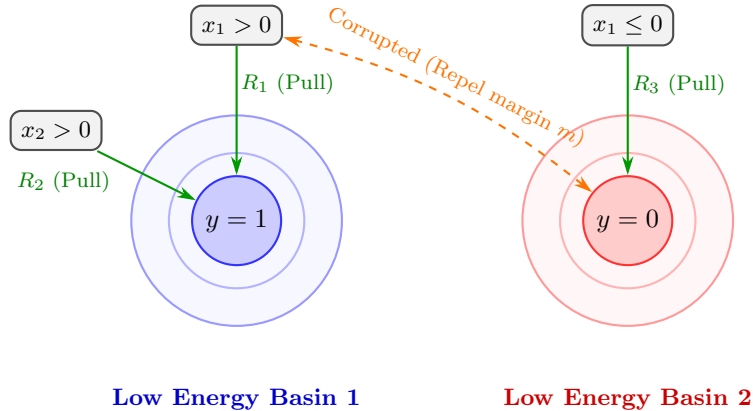
\begin{figure}[H]
    \centering
    \begin{tikzpicture}[
        node distance=1.5cm,
        basin_blue/.style={circle, draw=blue!40, thick, fill=blue!3, minimum size=2.8cm},
        basin_red/.style={circle, draw=red!40, thick, fill=red!3, minimum size=2.8cm},
        pole_blue/.style={circle, draw=blue!80, fill=blue!20, thick, minimum size=0.6cm, font=\bfseries},
        pole_red/.style={circle, draw=red!80, fill=red!20, thick, minimum size=0.6cm, font=\bfseries},
        antecedent/.style={rectangle, draw=black!70, fill=gray!10, rounded corners, thick, inner sep=4pt, font=\small},
        pull/.style={-{Stealth[length=2.5mm, width=1.5mm]}, thick, draw=green!60!black},
        repel/.style={<->, >={Stealth[length=2.5mm, width=1.5mm]}, dashed, thick, draw=orange!90!red}
    ]

    % Energy Basins (Outer limits)
    \node[basin_blue] (basin1) at (-2.6, 0) {};
    \node[basin_red] (basin0) at (2.6, 0) {};

    % Energy Contours (Inner) to show "depth"
    \draw[blue!30, thick] (-2.6, 0) circle (0.9cm);
    \draw[blue!20, thick] (-2.6, 0) circle (0.45cm);

    \draw[red!30, thick] (2.6, 0) circle (0.9cm);
    \draw[red!20, thick] (2.6, 0) circle (0.45cm);

    % Poles (Consequents)
    \node[pole_blue] (y1) at (-2.6, 0) {$y=1$};
    \node[pole_red] (y0) at (2.6, 0) {$y=0$};

    % Antecedents
    \node[antecedent] (A1) at (-2.6, 2.6) {$x_1 > 0$};
    \node[antecedent] (A2) at (-5.0, 1.2) {$x_2 > 0$};
    \node[antecedent] (A3) at (2.6, 2.6) {$x_1 \leq 0$};

    % Pull Arrows (Valid Rules)
    \draw[pull] (A1) -- node[midway, right, font=\footnotesize, text=green!50!black, xshift=-2pt, yshift=10pt] {$R_1$ (Pull)} (y1);
    \draw[pull] (A2) -- node[midway, above left, font=\footnotesize, text=green!50!black, xshift=-10pt, yshift=-10pt] {$R_2$ (Pull)} (y1);
    \draw[pull] (A3) -- node[midway, right, font=\footnotesize, text=green!50!black, xshift=-2pt, yshift=10pt] {$R_3$ (Pull)} (y0);

    % Repel Arrow (Invalid Rule) - Now Bi-directional
    \draw[repel] (A1) to[bend left=12] node[midway, above, sloped, font=\footnotesize, text=orange!90!red] {Corrupted (Repel margin $m$)} (y0);

    % Labels
    \node[font=\bfseries\small, text=blue!80!black, below=1.5cm of y1] {Low Energy Basin 1};
    \node[font=\bfseries\small, text=red!80!black, below=1.5cm of y0] {Low Energy Basin 2};

    \end{tikzpicture}
    \caption{Geometric illustration of the Energy-Based Constraints (EBC). Valid rule antecedents (gray boxes) are pulled into stable, low-energy basins corresponding to their correct consequents. Corrupted rule pairings are actively repelled into high-energy regions via the contrastive margin.}
    \label{fig:analytical_ebc}
\end{figure}

Under Energy-Based Constraint (EBC) training, as geometrically illustrated in Fig.\ref{fig:analytical_ebc}:
\begin{itemize}
    \item Valid rule pairs are pulled together. Antecedents sharing the same consequent (e.g., $x_1 > 0$ and $x_2 > 0$) are pulled towards the shared $y=1$ semantic pole, forming a single, stable low-energy basin. 
    \item Conversely, $R_3$ is pulled towards the $y=0$ pole, establishing a \textit{distinct}, geometrically separated low-energy basin for the alternate outcome.
    \item Corrupted rules (e.g., $x_1 > 0 \rightarrow y = 0$) act as negative samples and are actively repelled from these valid poles into high-energy regions via the contrastive margin.
\end{itemize}

This dynamic explicitly carves out separated energy valleys corresponding strictly to valid logical structures, preventing the model from interpolating invalid mappings between opposing semantic states.

\subsection{Simulation: A 3D Energy Landscape} \label{subsec:simulated_example}

To empirically validate this manifold shaping, we design a synthetic experiment mapping rule embeddings into a continuous 3D latent space. We explicitly control the ground-truth data generation process (DGP) to observe how rule supervision alters the learned geometry compared to pure data-driven learning. Simulation details are to be found in Appendix.\ref{sec:simulation_details}.

\paragraph{Data Generation Process (DGP)}
We simulate an environment where the underlying logic is governed by distinct rule families. The generation proceeds as follows:
\begin{enumerate}
    \item \textit{Latent Rule Sampling:} we sample ``ground truth'' rule embeddings $z_r^*$ from two distinct multivariate Gaussian distributions in $\mathbb{R}^3$, $\mathcal{N}(\mu_1, \Sigma_1)$ and $\mathcal{N}(\mu_2, \Sigma_2)$. These represent the latent geometric centers of our valid logical rule families.
    \item \textit{Rule Decoding:} each sampled embedding $z_r^*$ is projected onto a discrete symbolic rule structure $r = (A \rightarrow C)$, representing specific numeric transformations and target outcomes.
    \item \textit{Raw Data Generation:} we generate a dataset of raw observations $\mathcal{D}_{raw} = \{(x_i, y_i)\}_{i=1}^N$ that are strictly coherent with the decoded symbolic rules $r$. The features $x_i$ are sampled, and the labels $y_i$ are assigned deterministically by applying the rules. Additive observation noise is applied.
    \item \textit{Negative Rule Generation:} to establish a contrastive background, we generate invalid rules (negative pairs). Instead of uniform sampling, we draw antecedent embeddings $z_c$ from two distinct out-of-distribution (OOD) multivariate Gaussians, $\mathcal{N}(\mu_{neg1}, \Sigma_{neg1})$ and $\mathcal{N}(\mu_{neg2}, \Sigma_{neg2})$. Their corresponding consequents $C_{\text{neg}}$ are sampled randomly, ensuring they represent logical mappings $(A, C_{\text{neg}})$ that structurally contradict the ground-truth DGP.
\end{enumerate}

\paragraph{Experimental Setup}
We train two distinct predictive architectures to map the input context to the target space. Notably, because the raw data and symbolic rules in this controlled simulation exist within the exact same dimensionality and semantic scale, this experiment represents \textit{a special case} where the rule encoder and data encoder are unified into a single shared encoder network.
\begin{enumerate}
    \item \textit{Classic JEPA:} trained purely on raw data prediction using the standard self-supervised prediction loss over the data pairs:
    $$
    \mathcal{L}_{JEPA} = \mathbb{E}_{(x,y) \sim \mathcal{D}_{raw}} \left[ \|g(f_c(x)) - f_t(y)\|_2^2 \right]
    $$
    \item \textit{RiJEPA (EBC):} trained with the joint objective, combining raw data prediction with the explicit rule-based Energy-Based Constraint (Eq.\ref{eq:RiJEPA_loss}):
    $$
    \mathcal{L}_{total} = \mathcal{L}_{JEPA} + \beta \mathcal{L}_{EBC}
    $$
    where $\mathcal{L}_{EBC}$ explicitly enforces the rule validity boundary via (Eq.\ref{eq:RiJEPA_EBC_loss}):
    $$
    \mathcal{L}_{EBC} = \sum_{\text{valid}} E(A, C) + \lambda \sum_{\text{invalid}} \max(0, m - E(A, C_{\text{neg}}))
    $$
    and $E(A, C) = \|g(f_c(A)) - f_t(C)\|_2^2$ is the predicted energy of the embedded symbolic rule.
\end{enumerate}

\begin{figure}[H]
    \centering
    \includegraphics[width=0.85\textwidth]{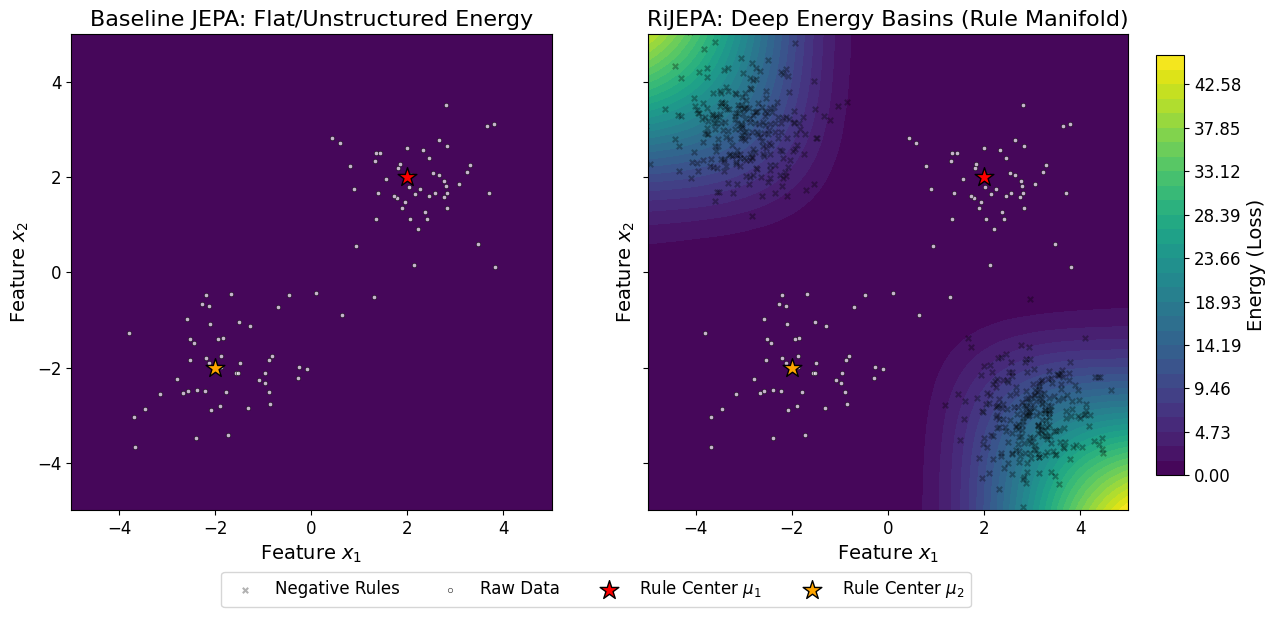}
    \caption{Energy landscapes of Classic JEPA and RiJEPA over the feature space. While the baseline learns an unstructured, flat functional mapping, RiJEPA carves deep, logically valid energy basins while actively repelling negative OOD rules.}
    \label{fig:energy_landscape}
\end{figure}

\paragraph{Results and Manifold Visualization}
We evaluate the models by computing the energy landscape $E(A, C)$ over the entire latent space, as visualized in Fig.\ref{fig:energy_landscape}. 

In the \textbf{Classic JEPA}, the learned latent space is dictated solely by raw data correlations. Because it never explicitly models the symbolic boundary of the underlying rules, the resulting energy landscape over the rule space is extremely flat and unstructured, as shown in the left panel of Fig.\ref{fig:energy_landscape}. The baseline model assumes the mapping is globally applicable, failing to establish a clear decision boundary for logical validity and leaving it highly susceptible to shortcut learning and OOD failures.

Conversely, \textbf{RiJEPA} explicitly sculpts the 3D manifold. By minimizing the energy of the valid rules and actively maximizing the energy of the clustered negative samples via the margin $m$, the EBC loss (Eq.\ref{eq:RiJEPA_EBC_loss}) carves deep, well-defined energy basins exactly centered at $\mu_1$ and $\mu_2$, as observed in the right panel of Fig.\ref{fig:energy_landscape}. Concurrently, the negative sample clusters are forced into steep high-energy peaks, creating a highly separable topological landscape. This confirms that rule supervision successfully injects a \textit{structural inductive bias}, transforming a purely correlational, entangled latent space into a robust, rule-aware energy manifold capable of precise geometric logic.

\paragraph{Quantitative Evaluation and Latent Geometry}

To rigorously evaluate the model's robustness to OOD hallucinations, we generated two dedicated test sets: an In-Distribution (ID) set sampled near the valid rule centers ($\mu_1, \mu_2$), and an OOD set sampled near the negative rule centers ($\mu_{neg1}, \mu_{neg2}$). We applied the true underlying rule transformation to both sets and evaluated the assigned energy. 

\begin{table}[H]
    \centering
    \caption{Mean Energy Assigned to the Rule Mapping.}
    \begin{tabular}{lcc}
        \hline
        \textbf{Model} & \textbf{ID Energy (Valid)} & \textbf{OOD Energy (Invalid)} \\
        \hline
        Classic JEPA & 0.003 & 0.150 \\
        RiJEPA (EBC) & 0.006 & 12.713 \\
        \hline
    \end{tabular}
    \label{tab:energy_results}
\end{table}

\begin{figure}[H]
    \centering
    \includegraphics[width=0.5\textwidth]{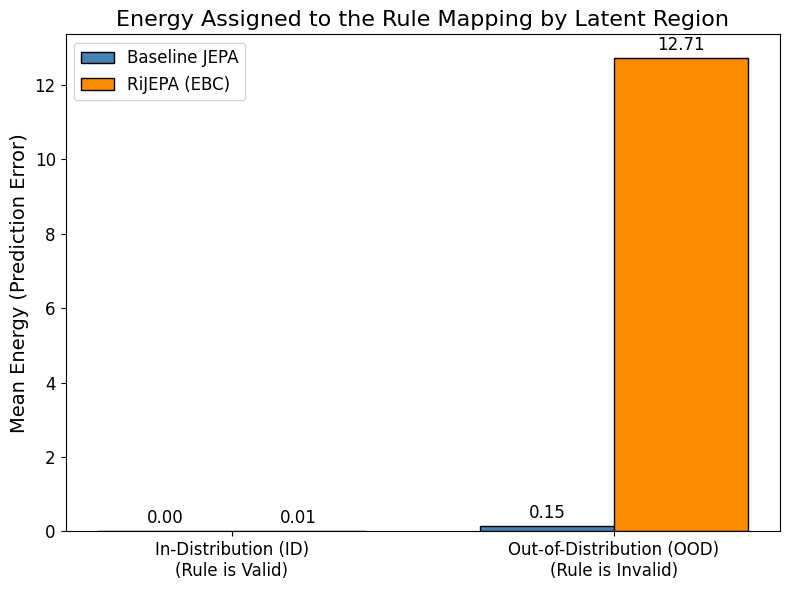} 
    \caption{Mean energy assignment by latent region. RiJEPA successfully applies a massive energy penalty to invalid rules outside the data distribution.}
    \label{fig:bar_chart}
\end{figure}

As shown in Table.\ref{tab:energy_results} and Fig.\ref{fig:bar_chart}, the Classic JEPA erroneously assigns extremely low energy to the OOD data ($\approx 0.150$), demonstrating a dangerous over-extrapolation of the rule into invalid regions. Conversely, RiJEPA perfectly discriminates the regions, maintaining low energy for valid data ($\approx 0.006$) while assigning a massive energy penalty ($\approx 12.713$) to the OOD hallucinations.

\begin{figure}[H]
    \centering
    \includegraphics[width=0.85\textwidth]{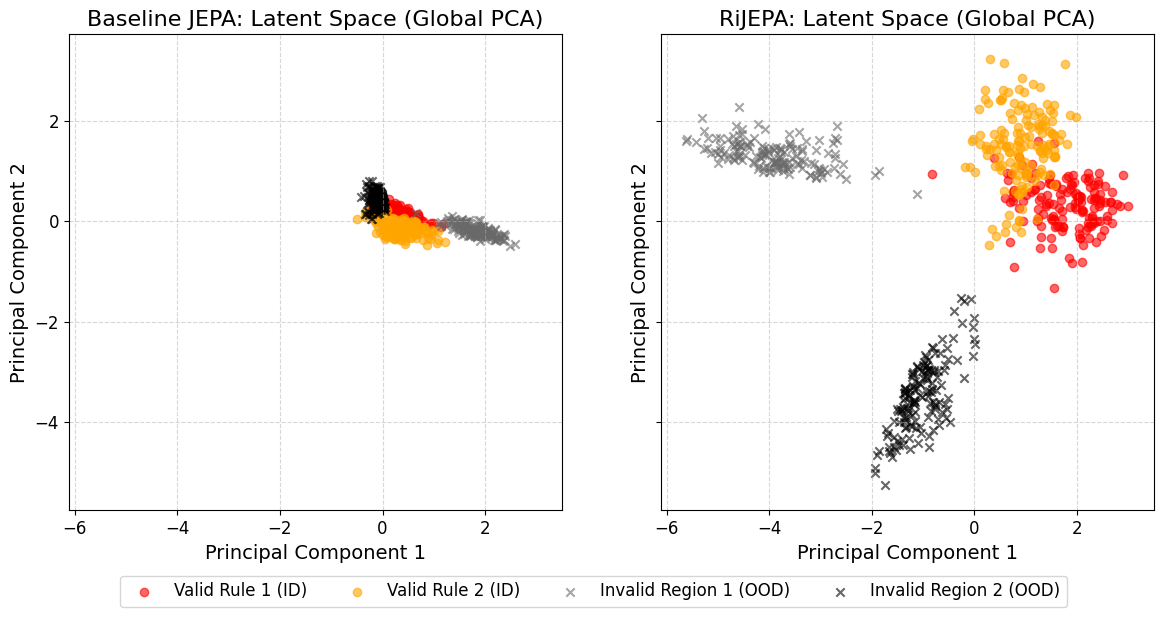}
    \caption{Global PCA projection of latent representations. RiJEPA actively clusters valid rules and strictly isolates invalid regions, whereas the Classic JEPA leaves the semantic space entangled.}
    \label{fig:pca_latent}
\end{figure}

To understand the internal mechanism driving this behavior, we visualize the learned latent representations ($z_c$) using \textit{Principal Component Analysis} (PCA) in Fig.\ref{fig:pca_latent}. To ensure mathematically rigorous comparability, we fit a single, universal PCA projection across the combined embeddings of both models, sharing the exact same spatial axes. It is observed that, in the Classic JEPA, the latent embeddings of valid rules (circles) and invalid OOD regions (crosses) are deeply entangled, preventing the predictive module from establishing a logical boundary. However, in RiJEPA, the Energy-Based Constraint actively structures the latent geometry. Valid rules are tightly clustered and explicitly isolated from the invalid rule regions. This proves our core hypothesis: rule supervision does not merely adjust the final output mapping; it explicitly organizes the internal latent space to reflect structural, compositional logic, providing a robust, interpretable foundation for downstream continuous rule discovery.

\subsection{Real-world Case Study: Clinical Rule Discovery and Neural-Symbolic Alignment} \label{subsec:clinical_case}

To validate the RiJEPA framework in a high-stakes clinical domain, we applied the architecture to the UCI Heart Disease dataset\footnote{The heart risk dataset is available at: \url{https://archive.ics.uci.edu/dataset/45/heart+disease}, we used the processed \textit{Cleveland data}.} \cite{DETRANO1989304}. This experiment serves as a practical demonstration of how the model bridges the gap between raw clinical observations and symbolic medical knowledge. Unlike standard predictive models, RiJEPA is evaluated on its ability to organize a latent manifold that obeys the foundational laws of clinical cardiology.

\paragraph{Data}
The dataset comprises clinical and non-invasive test results originally collected from 303 consecutive patients referred for coronary angiography at the Cleveland Clinic \cite{DETRANO1989304}. While the complete heart disease database includes 76 attributes, in this experiment we utilize the standard processed subset of 14 clinical and test variables most pertinent to the diagnosis of angiographic coronary disease. These include four primary clinical variables: age, sex, chest pain type (typical anginal, atypical anginal, non-anginal, or asymptomatic), and resting systolic blood pressure. The remaining attributes consist of nine routine or non-invasive test results:
\begin{itemize}
    \item Serum cholesterol and fasting blood sugar levels.
    \item Resting electrocardiographic (ECG) results, categorized by ST-T wave abnormalities or left ventricular hypertrophy.
    \item Exercise-induced markers, including maximum heart rate achieved, the presence of exercise-induced angina, and the peak exercise ST-segment slope.
    \item Fluoroscopic data indicating the number of major vessels containing calcium.
\end{itemize}
The target attribute, \textit{num} (of type integer), represents the angiographic disease status (diagnosis of heart disease). This dataset provides a robust foundation for neural-symbolic alignment, as it allows the model to reconcile high-dimensional patient observations with the established logical constraints of clinical cardiology.

To explicitly define the operational semantics of the dataset, Table.\ref{tab:variables} provides a comprehensive breakdown of the 14 attributes utilized, including their data types and clinical encoding schemas.

\begin{table}[H]
\centering
\caption{Summary of the 14 Clinical and Non-invasive Test Variables}
\label{tab:variables}
\resizebox{\textwidth}{!}{
\begin{tabular}{lll}
\hline
\textbf{Variable} & \textbf{Type} & \textbf{Description \& Encoding Schema} \\ \hline
\texttt{age} & Integer & Patient age in years \\
\texttt{sex} & Categorical & Patient sex ($1 = \text{male}$, $0 = \text{female}$) \\
\texttt{cp} & Categorical & Chest pain type ($1: \text{typical angina}, 2: \text{atypical angina}, 3: \text{non-anginal}, 4: \text{asymptomatic}$) \\
\texttt{trestbps} & Integer & Resting blood pressure on hospital admission (in mm Hg) \\
\texttt{chol} & Integer & Serum cholesterol measurement (in mg/dl) \\
\texttt{fbs} & Categorical & Fasting blood sugar $> 120$ mg/dl ($1 = \text{true}$, $0 = \text{false}$) \\
\texttt{restecg} & Categorical & Resting ECG results ($0: \text{normal}, 1: \text{ST-T wave abnormality}, 2: \text{left ventricular hypertrophy}$) \\
\texttt{thalach} & Integer & Maximum heart rate achieved during exercise testing \\
\texttt{exang} & Categorical & Exercise-induced angina observed ($1 = \text{yes}$, $0 = \text{no}$) \\
\texttt{oldpeak} & Continuous & ST-segment depression induced by exercise relative to rest \\
\texttt{slope} & Categorical & Slope of the peak exercise ST segment ($1: \text{upsloping}, 2: \text{flat}, 3: \text{downsloping}$) \\
\texttt{ca} & Integer & Number of major vessels ($0-3$) colored by fluoroscopy \\
\texttt{thal} & Categorical & Thalassemia diagnosis ($3: \text{normal}, 6: \text{fixed defect}, 7: \text{reversible defect}$) \\
\texttt{target\_risk} & Categorical & Angiographic disease status ($0: < 50\% \text{ diameter narrowing}, 1: \ge 50\% \text{ diameter narrowing}$) \\ \hline
\end{tabular}}
\end{table}

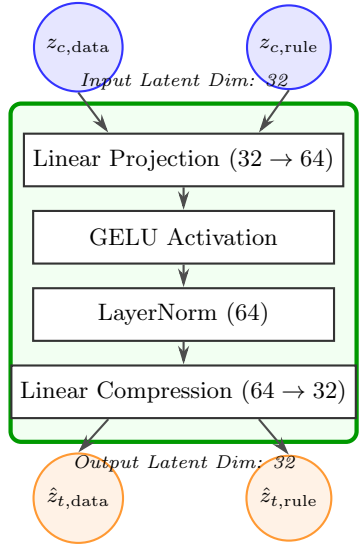
\begin{figure}[H]
    \centering
    \begin{tikzpicture}[
        node distance=0.8cm,
        layer/.style={rectangle, draw=black!80, fill=white, thick,
                      text centered, minimum width=4.0cm, minimum height=0.7cm,
                      font=\small},
        shared/.style={rectangle, draw=green!60!black, fill=green!5,
                       ultra thick, rounded corners,
                       minimum width=4.6cm, minimum height=4.5cm},
        input_node/.style={circle, draw=blue!80, fill=blue!10,
                           thick, minimum size=0.8cm, font=\footnotesize},
        output_node/.style={circle, draw=orange!80, fill=orange!10,
                            thick, minimum size=0.8cm, font=\footnotesize},
        arrow/.style={-{Stealth[length=2.5mm, width=1.5mm]}, thick, draw=black!70}
    ]

    % ---------- SHARED PREDICTOR BOX (All Layers Inside) ----------
    \node[shared] (g_box) at (0, -1.5) {};

    % ---------- INTERNAL LAYERS (Centered inside g_box) ----------
    \node[layer] (lin1) at (0, 0)
        {Linear Projection ($32 \rightarrow 64$)};

    \node[layer, below=0.3cm of lin1] (gelu)
        {GELU Activation};

    \node[layer, below=0.3cm of gelu] (norm)
        {LayerNorm ($64$)};

    \node[layer, below=0.3cm of norm] (lin2)
        {Linear Compression ($64 \rightarrow 32$)};

    % ---------- INPUTS (Positioned just above the box) ----------
    \node (input_start) at (0, 1.5) {}; 
    \node[input_node] (z_c_data) at (-1.4, 1.5) {$z_{c,\mathrm{data}}$};
    \node[input_node] (z_c_rule) at (1.4, 1.5) {$z_{c,\mathrm{rule}}$};

    \node[below=0.1cm of input_start, font=\scriptsize\itshape]
        {Input Latent Dim: 32};

    % ---------- OUTPUTS (Positioned just below the box) ----------
    \node (output_anchor) at (0, -4.5) {};
    \node[output_node] (z_t_data) at (-1.4, -4.5) {$\hat z_{t,\mathrm{data}}$};
    \node[output_node] (z_t_rule) at (1.4, -4.5) {$\hat z_{t,\mathrm{rule}}$};

    \node[above=0.1cm of output_anchor, font=\scriptsize\itshape]
        {Output Latent Dim: 32};

    % ---------- INTERNAL FLOW ----------
    \draw[arrow] (lin1) -- (gelu);
    \draw[arrow] (gelu) -- (norm);
    \draw[arrow] (norm) -- (lin2);

    % ---------- INPUT TO LAYERS ----------
    \draw[arrow] (z_c_data.south) -- (lin1.160); 
    \draw[arrow] (z_c_rule.south) -- (lin1.20);

    % ---------- LAYERS TO OUTPUT ----------
    \draw[arrow] (lin2.200) -- (z_t_data.north);
    \draw[arrow] (lin2.340) -- (z_t_rule.north);

    \end{tikzpicture}
    \caption{The universal predictor $g(\cdot)$. The predictor facilitates modality-agnostic reasoning by utilizing a standardized latent bottleneck of $d=32$. It accepts either continuous data context embeddings ($z_{c\_data}$) or symbolic rule antecedent embeddings ($z_{c\_rule}$), processing them through a shared non-linear transformation to yield their respective target predictions.}
    \label{fig:clinical_predictor_architecture}
\end{figure}

\paragraph{Architectural Designs}
We use the Multi-Modal Dual-Encoder RiJEPA architecture in Fig.\ref{fig:clinical_predictor_architecture} to project fundamentally different data types (continuous patient vitals and discrete symbolic rules) into a Shared Latent Semantic Space. The system employs separate encoders for each modality to handle their distinct dimensionality and semantic scale. 

The shared Predictor ($g$), shown in Fig.\ref{fig:clinical_predictor_architecture}, is the core neuro-symbolic bridge. It is a non-linear projection head ($\mathbb{R}^{32} \to \mathbb{R}^{32}$) designed to map a context latent to a target latent through an expansion ($32 \rightarrow 64$) and compression ($64 \rightarrow 32$) cycle stabilized by Layer Normalization. As the latent dimension is standardized ($d=32$) across all encoders, $g$ acts as a modality-agnostic "Shared Semantic Bridge". During training, $g$ learns a unified inference logic by processing three distinct data paths: 
\begin{itemize}
    \item \textit{Data Path:} mapping vitals to future states ($g(z_{c,data}) \approx z_{t,data}$).
    \item \textit{Rule Path:} mapping antecedents to consequents ($g(z_{c,rule}) \approx z_{t,rule}$).
    \item \textit{Anchor Path:} aligning vitals with symbolic clinical outcomes.
\end{itemize}
In our implementation, the predictor $g$ accepts the data embeddings and rule embeddings separately, not simultaneously (i.e., they are never concatenated). Because $g$ functions as a universal projection head mapping from $\mathbb{R}^{32} \rightarrow \mathbb{R}^{32}$, it is called multiple times independently for different modalities during a single training step:
\begin{enumerate}
    \item First, it processes the \textit{Data Context}, accepting only the 32-dimensional raw data embedding.
    \item Next, it processes the \textit{Valid Rule Antecedents}, accepting only the 32-dimensional valid symbolic rule embedding.
    \item Finally, it processes the \textit{Invalid Rule Antecedents}, accepting only the 32-dimensional invalid or corrupted symbolic rule embedding.
\end{enumerate}
Processing these modalities separately is a critical architectural decision. If the embeddings were concatenated (e.g., yielding a 64-dimensional input), the predictor would learn to expect both modalities to be present at all times to make a prediction. By passing them independently through the exact same shared weights, $g$ is forced to become fully \textit{modality-agnostic}. It learns that a 32-dimensional vector representing a patient's high blood pressure from the data encoder ($f_{c\_data}$) and a 32-dimensional vector representing the text string \texttt{trestbps\_level=High} from the rule encoder ($f_{c\_rule}$) must be transformed using the exact same mathematical pathway.

All of these independent forward passes generate separate loss terms (i.e., $\mathcal{L}_{JEPA}$, $\mathcal{L}_{EBC}$, and an explicit anchor loss $\mathcal{L}_{anchor}$), which are then linearly combined (adding an anchor loss term to Eq.\ref{eq:RiJEPA_loss}):
\begin{equation} \label{eq:RiJEPA_loss_withAnchorLoss}
\mathcal{L}_{total} = \mathcal{L}_{JEPA} + \alpha\mathcal{L}_{EBC} + \beta\mathcal{L}_{anchor}
\end{equation}
with weight hyper-parameters $\alpha=2.0$, $\beta=5.0$, and individual loss components defined mathematically as follows:
\begin{align}
\mathcal{L}_{JEPA} &= \mathbb{E}_{x} \left[ \| g(f_{c\_data}(x_c)) - f_{t\_data}(x_t) \|_2^2 \right] \label{eq:jepa_loss} \\
\mathcal{L}_{EBC} &= \mathbb{E}_{A_p, C_p} \left[ \| z_{hp} - z_{tp} \|_2^2 \right] + \mathbb{E}_{A_n, C_n} \left[ \max(0, m - \| z_{hn} - z_{tn} \|_2^2) \right] \label{eq:ebc_loss} \\
\mathcal{L}_{anchor} &= \mathbb{E}_{x, y} \left[ y \| \hat{z}_{t,data} - z_{rh} \|_2^2 + (1-y) \| \hat{z}_{t,data} - z_{rl} \|_2^2 \right] \label{eq:anchor_loss}
\end{align}
Here, the Energy-Based Constraint ($\mathcal{L}_{EBC}$) applies a contrastive margin $m=2.0$ to push apart the predicted states ($z_{hn}$) and target consequents ($z_{tn}$) for invalid or corrupted rules ($A_n, C_n$), while pulling together valid rule pairs ($z_{hp}, z_{tp}$). The Anchor Loss ($\mathcal{L}_{\text{anchor}}$) acts as an \emph{absolute centroid alignment}. We use predefined, fixed symbolic poles ($z_{rh}$ for high risk, $z_{rl}$ for low risk) mapped by $f_{t\_rule}$, and penalize the absolute distance between a patient's predicted continuous state ($\hat{z}_{t,data} = g(f_{c\_data}(x_c))$) and these fixed coordinates based on their true diagnostic outcome $y \in \{0,1\}$. ``Anchor'' here therefore means ``tying down'' to a fixed semantic point.

When backpropagation is executed on this combined loss, the gradient contributions from all independent passes are accumulated through $g$, updating its weights to satisfy the continuous data constraints and the discrete symbolic rule constraints simultaneously. Rules are processed as "Neuro-Symbolic" forward passes, acting as geometric constraints rather than standard training inputs. The encoders $f_{c\_rule}$ and $f_{t\_rule}$ are updated via standard backpropagation, allowing the model to learn symbolic representations. In contrast, the data target encoder $f_{t\_data}$ remains frozen and its parameters ($\theta_{t\_data}$) are updated via an Exponential Moving Average (EMA) from the context encoder parameters ($\theta_{c\_data}$) \cite{assran2023self}:
\begin{equation} \label{eq:ema_update}
\theta_{t\_data} \leftarrow \tau \theta_{t\_data} + (1 - \tau) \theta_{c\_data}
\end{equation}
governed by a momentum parameter of $\tau = 0.99$ to ensure representation stability. This configuration ensures that clinical data and symbolic logic end up in the same geometric location within the manifold. Detailed layer dimensions and activation functions are provided in Appendix~\ref{app:real_world_exp_details}.

\paragraph{Rule Extraction}
To generate the necessary symbolic knowledge base for the RiJEPA architecture, we applied frequent itemset mining to the training split. Continuous clinical variables (e.g., age, cholesterol, maximum heart rate) were first discretized into clinically meaningful categorical bins (e.g., \texttt{age\_group}, \texttt{chol\_level}, \texttt{thalach\_level}) to facilitate logical abstraction. The processed dataset was subsequently transformed into a transactional format, and the FP-Growth algorithm \cite{han2000mining} was applied to mine association rules. We enforced a minimum support threshold of $0.04$ (4\% occurrence in the dataset) and a minimum confidence threshold of $0.70$ (70\% predictive certainty).

By filtering the mined rules to strictly retain those projecting a consequent of either \texttt{target\_risk=1.0} (High Risk) or \texttt{target\_risk=0.0} (Low Risk), the system yielded a robust logical prior of 4801 distinct rules. Table.\ref{tab:rule_examples} illustrates a selection of these extracted rules, demonstrating both protective physiological profiles (yielding low risk) and severe diagnostic markers (yielding high risk).

\begin{table}[H]
\centering
\scriptsize
\setlength{\tabcolsep}{4pt}   % reduce column spacing (default = 6pt)
\caption{Examples of Extracted Clinical Rules via FP-Growth}
\label{tab:rule_examples}
\begin{tabular}{l p{7.5cm} c c}
\hline
\textbf{Risk Profile} &
\textbf{Symbolic Logic (Antecedent $\rightarrow$ Consequent)} &
\textbf{Support} &
\textbf{Confidence} \\
\hline
Low Risk &
\texttt{age\_group=Middle} AND \texttt{thalach\_level=High}
$\rightarrow$ \texttt{target\_risk=0.0} &
21.5\% & 70.8\% \\
Low Risk &
\texttt{cp=1.0} AND \texttt{restecg=2.0}
$\rightarrow$ \texttt{target\_risk=0.0} &
4.2\% & 83.3\% \\
High Risk &
\texttt{cp=4.0} AND \texttt{slope=3.0}
$\rightarrow$ \texttt{target\_risk=1.0} &
4.2\% & 100.0\% \\
\hline
\end{tabular}
\end{table}

\paragraph{Quantitative Performance and Manifold Alignment}
The fundamental advantage of RiJEPA is that the symbolic rules act strictly as a structural inductive bias during training, explicitly shaping the latent manifold. At test time, the rule-mining algorithm (e.g., FP-Growth) is completely turned off, and the symbolic rule context pathways ($f_{c\_rule}$) are not used to process the test data. Once the network is fully trained, the data encoder ($f_{c\_data}$) and predictor ($g$) have learned how to map raw patient data directly into this logically organized space independently. The system operates strictly as a feed-forward pipeline: New Patient Data ($x_{\text{test}}$) $\rightarrow f_{c\_data} \rightarrow g \rightarrow$ Compare to predefined logical boundaries. This makes inference extremely fast, deterministic, and highly interpretable, as every prediction can be plotted and visually justified.

To rigorously evaluate this capacity to distill raw clinical observations into a logically sound representation space, we designed a comprehensive evaluation suite to assess both the predictive power and the structural integrity of the learned manifolds. First, \textit{Downstream Linear Probing} evaluates the inherent linear separability of the learned features by fitting a simple linear classifier (logistic regression) on top of the frozen continuous data embeddings ($f_{c\_data}(x)$). Second, \textit{Zero-Shot Logical Inference} tests the core neural-symbolic alignment: it classifies unseen test patients without any task-specific fine-tuning by simply computing the Euclidean distance between the patient's predicted state ($g(z_{c\_data})$) and the explicit symbolic poles ($f_{t\_rule}(C)$) for ``High Risk'' and ``Low Risk''. Third, a \textit{Fallback Latent Norm} ablation test measures the architectural stability of the network when symbolic rule inputs are completely missing or masked (i.e., feeding a zero-tensor to the rule encoder). All tasks were evaluated on the separated, unseen test set.

For the \textit{Downstream Linear Probing} and manifold visualization tasks, we also introduced a \textbf{Classic JEPA} baseline for fair comparison. The Classic JEPA processes the exact same raw clinical data but utilizes only the data encoders ($f_{c\_data}, f_{t\_data}$) and predictor ($g$). It is trained purely using the standard (latent) data-to-data prediction loss ($\mathcal{L}_{JEPA}$) and the EMA target encoder update, without any symbolic rule inputs, Energy-Based Constraints, or explicit anchor losses. RiJEPA, on the other hand, was trained on both raw data and rules extracted from it; at test time, it performs predictions requiring only the raw test data ($x_{\text{test}}$).

\begin{figure}[H]
    \centering
    \includegraphics[width=0.75\textwidth]{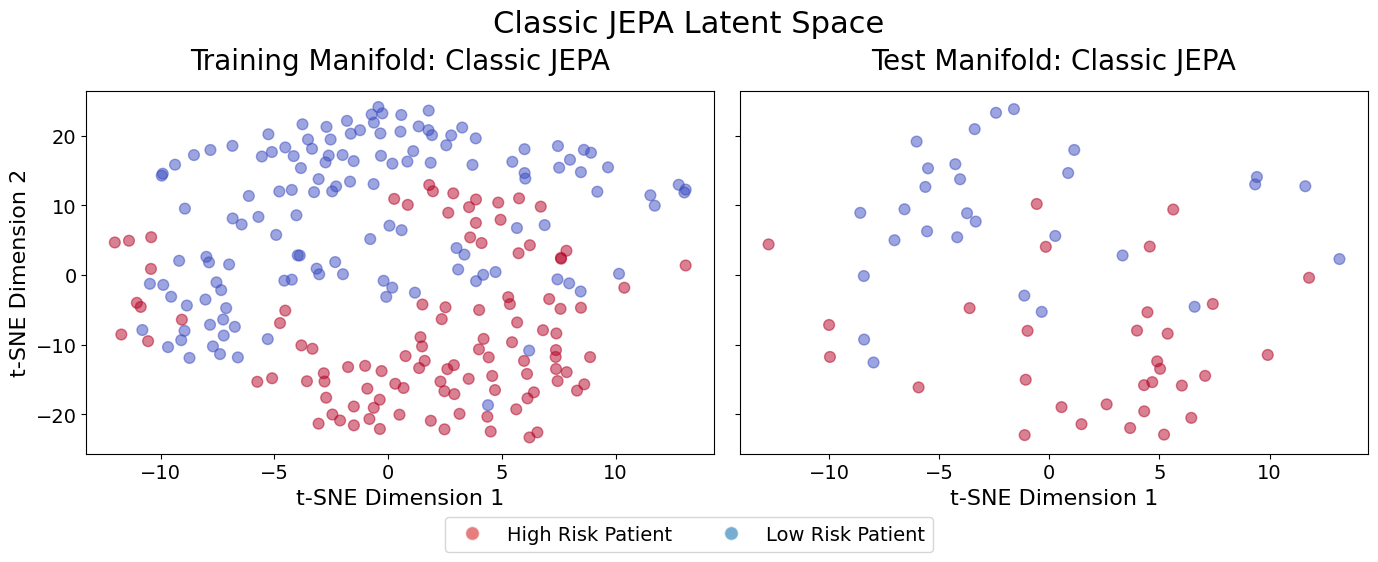} \\
    \includegraphics[width=0.75\textwidth]{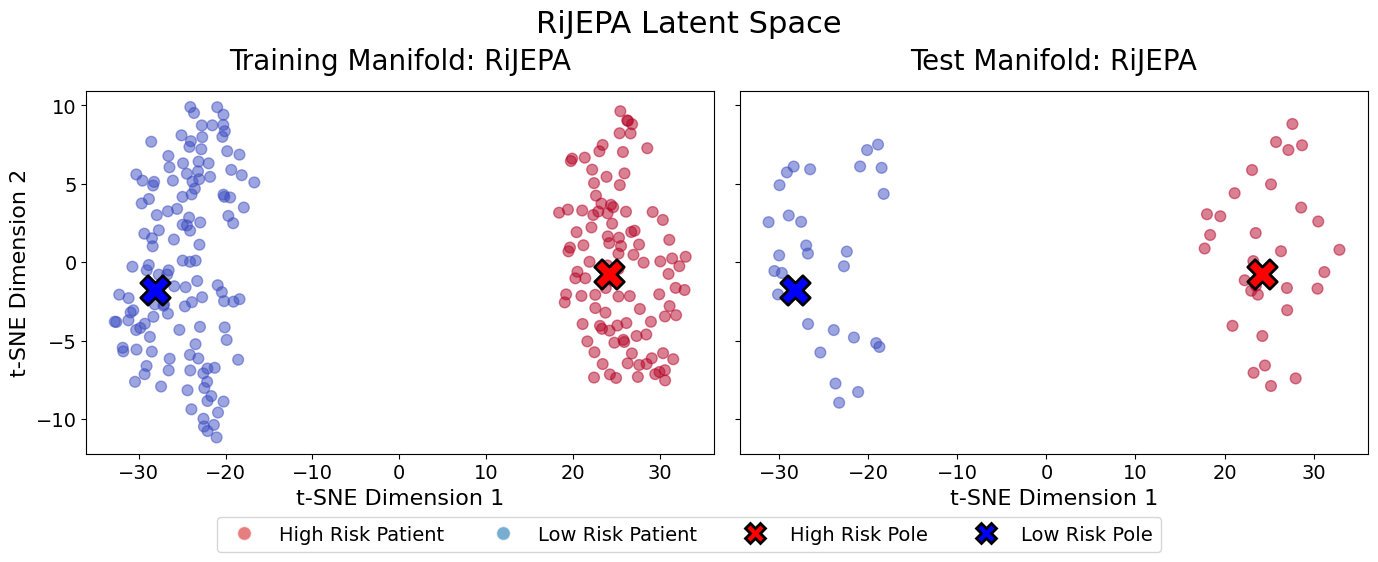}
    \caption{t-SNE visualization comparing the Classic JEPA (upper) and RiJEPA (bottom) latent spaces. While Classic JEPA yields an unstructured, continuous manifold where risk profiles are visually entangled, RiJEPA's Energy-Based Constraints explicitly pull continuous patient data (circles) into highly separable clusters strictly anchored around discrete symbolic poles (crosses).}
    \label{fig:manifolds}
\end{figure}

The impact of the Energy-Based Constraints (EBC) on the representation space is most immediately apparent when visualizing its topology. It is crucial to distinguish between high-dimensional linear separability and topological clustering. As visualized in the t-SNE latent manifold plots (Fig.\ref{fig:manifolds}), the Classic JEPA produces a highly entangled, continuous manifold. Because it lacks an explicit diagnostic boundary or margin constraint, patients transition from low to high risk along a smooth, unclustered gradient. In stark contrast, the EBC actively shatters this continuous gradient; it explicitly organizes RiJEPA's shared semantic space, pulling continuous patient data (circles) into highly separable, discrete clusters strictly anchored around their corresponding symbolic poles (crosses). 

These geometric observations translate directly into the empirical performance summarized in Table.\ref{tab:clinical_metrics}. The results demonstrate the unique advantages of the neuro-symbolic bridge. Due to the robust representation power of the underlying self-supervised JEPA backbone, both models project the data into a 32-dimensional space where a downstream linear hyper-plane can successfully slice the gradient (both achieving 100\% on Downstream Linear Probing). However, the Classic JEPA fundamentally relies on training this \textit{ex-post} supervised classifier to draw that arbitrary boundary and make diagnostic sense of the features. In stark contrast, RiJEPA intrinsically aligns and separates its representation space based on diagnostic logic during pretraining. This allows it to achieve perfect Data-to-Rule alignment (100\% Zero-Shot Logical Accuracy) and instantly classify patients based purely on geometric proximity to the semantic poles, completely eliminating the need for downstream supervised fine-tuning.

\begin{table}[H]
\centering
\scriptsize
\setlength{\tabcolsep}{4pt}
\renewcommand{\arraystretch}{1.2}
\caption{Quantitative Evaluation on the UCI Heart Risk Test Set}
\label{tab:clinical_metrics}
\begin{tabular}{p{3.8cm} c c p{5.5cm}}
\hline
\textbf{Evaluation Metric} & \textbf{Classic JEPA} & \textbf{RiJEPA} & \textbf{Clinical Significance} \\
\hline
Downstream Linear Probing & 100\% & 100\% & Both models allow for ex-post high-dimensional linear separation. \\
Zero-Shot Logical Accuracy & NA & 100\% & Only RiJEPA forms explicit logical clusters, enabling zero-shot diagnosis. \\
Fallback Latent Norm (Ablation) & NA & 5.87 & RiJEPA maintains geometric stability even without symbolic input. \\
\hline
\end{tabular}
\end{table}

The achieved Zero-Shot Logical Accuracy of 1.0 for RiJEPA is particularly significant. It mathematically confirms that the Energy-Based Constraints successfully warped the latent manifold, pulling patient data towards the correct symbolic "poles" purely through geometric alignment. Further, the Fallback Latent Norm remained robust at 5.87. Mathematically, this ablation test is executed by simulating a missing rule: we generate a zero-tensor matching the exact dimension of the rule vocabulary, which logically represents an empty symbolic antecedent ``IF [Absolutely Nothing]...''. This null input is processed through the rule context encoder ($f_{c\_rule}$). By calculating the standard $L_2$ norm of the predicted target vector ($z_{null}$), we prove that if the rule modality is completely missing or masked out during inference, the shared predictor handles the baseline geometry gracefully without numerically collapsing, diverging to infinity, or outputting \texttt{NaN} values.

\paragraph{The Logical Regularization Effect}
A critical observation during the training phase is the \textit{Logic-Accuracy Trade-off}. While a classic JEPA aims exclusively to minimize mean-squared error (MSE) on continuous data reconstruction by learning any available statistical correlation, RiJEPA accepts a slightly higher raw predictive error on the data to simultaneously satisfy absolute symbolic constraints. 

\begin{figure}[H]
    \centering
    \includegraphics[width=0.5\textwidth]{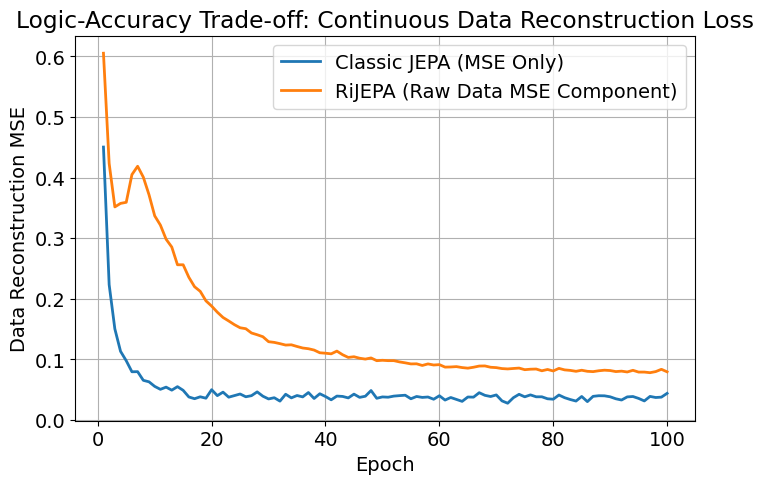}
    \caption{The Logic-Accuracy Trade-off. The Classic JEPA purely optimizes for continuous data correlation, achieving a lower raw reconstruction MSE. RiJEPA absorbs a slightly higher data prediction error in order to structurally satisfy the injected logical rule constraints.}
    \label{fig:mse_tradeoff}
\end{figure}

As visualized in Fig.\ref{fig:mse_tradeoff}, tracking the pure data-reconstruction component of the loss ($\mathcal{L}_{JEPA}$) reveals that Classic JEPA descends to a lower MSE floor. However, this "Logical Regularization" explicitly ensures that RiJEPA ignores "spurious" statistical shortcuts that might reconstruct the data efficiently but violate established clinical logic, yielding a more interpretatively sound representation space.

\paragraph{Generative Discovery and Abductive Reasoning}
Beyond purely discriminative tasks, a fundamental motivation for relaxing discrete logic into a continuous energy manifold is to enable bidirectional, generative reasoning. Notably, a Classic JEPA is mathematically \textit{incapable} of these tasks; because its target space maps strictly to continuous patient vitals without a corresponding rule encoder dictionary ($f_{c\_rule}$, $f_{t\_rule}$), it possesses no mechanism to translate geometric coordinates into interpretable, discrete clinical concepts. RiJEPA, by contrast, establishes a shared semantic energy surface ($E = \|g(z_c) - z_t\|_2^2$) grounded in symbolic logic. 

In real-world medical practice, experts do not merely predict outcomes from observations; they also hypothesize disease progression (forward inference), deduce probable underlying causes from observed outcomes (abductive reasoning), and translate raw physiological states into descriptive clinical profiles. To evaluate whether RiJEPA has successfully internalized this causal structure, we tested its generative capabilities using gradient-guided Langevin Dynamics. By iteratively updating latent vectors to minimize the structural energy of the shared predictor ($E = \|g(z_c) - z_t\|_2^2$), the model navigates the manifold to discover stable, logically valid clinical states.

We assessed RiJEPA's generative capacity across four distinct paradigms (Table~\ref{tab:langevin_results}):
\begin{enumerate}
    \item \textit{Joint Discovery:} operates unconditionally, allowing the system to freely wander the low-energy basins of the landscape to hallucinate entirely novel, valid clinical associations.
    \item \textit{Forward Inference:} fixes a specific antecedent condition (e.g., ``IF Senior Age...'') and generates the expected physiological and diagnostic consequences.
    \item \textit{Abductive Reasoning:} reverses the predictive flow by fixing a diagnostic conclusion (e.g., ``High Risk BECAUSE...'') and prompting the model to reconstruct the most probable physiological antecedents.
    \item \textit{Marginal-Predictive Discovery:} maps an actual, unseen patient's continuous physiological vitals ($x_{\text{test}}$) directly into the discrete rule-consequent space, translating a numerical state into a readable diagnostic profile.
\end{enumerate}

The empirical results confirm the model's fluency across all four paradigms. In the \textit{Joint Discovery} setting, the model synthesized a comprehensive clinical profile without external priors, associating normal cholesterol, elevated blood pressure, and left ventricular hypertrophy with severe outcomes like high diagnostic risk and downsloping ST segments. Interestingly, the unconstrained diffusion also hallucinated a dual-age profile (identifying both Young and Senior markers simultaneously), demonstrating the creative, interpolative nature of continuous energy manifolds when traversing boundary zones. 

During \textit{Forward Inference}, when constrained solely by the demographic prior of ``Senior Age'', the system logically projected a clinical trajectory culminating in high diagnostic risk, correctly predicting associated comorbidities such as medium heart rate, exercise angina, and left ventricular hypertrophy. The \textit{Abductive Reasoning} results represent powerful empirical evidence of clinical internalization. When prompted solely with the diagnostic conclusion of ``High Risk,'' the model successfully traced the outcome back to geometric root causes, reconstructing a profile containing severe, real-world markers such as exercise-induced angina ($exang=1.0$), low/medium heart rate, downsloping ST segments ($slope=3.0$), and senior age.

Finally, the \textit{Marginal-Predictive Discovery} task successfully demonstrated zero-shot clinical translation. By passing a specific test patient's continuous, standard-scaled vitals (characterized by young age and high maximum heart rate) through the data pathway ($g(f_{c\_data}(x_{\text{test}}))$) and decoding the nearest symbolic neighbors, the model accurately translated the raw measurements into a protective, low-risk clinical narrative: Low Risk, Normal ECG, Upsloping ST, High Maximum Heart Rate, and No Exercise Angina.

\begin{table}[H]
\centering
\scriptsize
\setlength{\tabcolsep}{4pt}
\renewcommand{\arraystretch}{1.2}
\caption{Generative Clinical Discovery through Langevin Diffusion Paradigms}
\label{tab:langevin_results}
\begin{tabular}{p{2.6cm} p{3.2cm} p{5.5cm}}
\hline
\textbf{Paradigm} & \textbf{Query Logic} & \textbf{Discovered Clinical Profile} \\ \hline
Joint Discovery & Unconditional & IF Normal Chol + Elevated BP + LV Hypertrophy + Downsloping ST + Typical Angina \newline $\implies$ High Risk + Downsloping ST + High FBS + (Mixed Age Markers) \\
\hline
Forward Inference & IF Senior THEN... & High Risk + Medium Heart Rate + Exercise Angina + Asymptomatic Pain + LV Hypertrophy \\
\hline
Abductive Reasoning & High Risk BECAUSE... & Exercise Angina + Low/Medium Heart Rate + Downsloping ST + Senior Age \\ 
\hline
Marginal-Predictive & \begin{tabular}[t]{@{}l@{}}Real Patient Vitals \\ (e.g., Young, High Max HR)\end{tabular} & Low Risk + Normal ECG + Upsloping ST + High Heart Rate + No Angina \\ \hline
\end{tabular}
\end{table}

\subsection{Summary: A Unified Semantic Space}

The empirical evidence presented in this real-world case study demonstrates that RiJEPA successfully converges the continuous data latent space and the discrete rule latent space into a unified \textit{Shared Semantic Space}. Visualizing this manifold as anchored to explicit symbolic poles fundamentally addresses the "Black Box" problem inherent to deep learning, offering a transparent, distance-based geometric justification for every diagnostic inference. 

Ultimately, while classic self-supervised architectures like JEPA are highly capable of projecting data into continuous, high-dimensional spaces that can be ex-post linearly separated, they fail to establish explicit structural boundaries. They fundamentally rely on training downstream supervised classifiers to draw decision boundaries across entangled manifolds. In contrast, RiJEPA achieves 100\% Zero-Shot logical accuracy simply by measuring the geometric distance to learned logical rules, completely eliminating the need for downstream supervision. Further, the model's fluency across generative joint discovery, forward inference, abductive reasoning, and marginal-predictive translation confirms that it is not merely a discriminative classifier. Rather, RiJEPA functions as a robust Neural-Symbolic Engine, capable of complex, bidirectional reasoning that seamlessly bridges the gap between raw neural pattern recognition and expert medical logic.

\section{Discussion} \label{sec:discussion}

At its core, this work addresses the fundamental challenge of knowledge representation. Real-world knowledge is traditionally captured in two distinct modalities: raw observational data and abstracted symbolic rules. Raw data is deeply comprehensive and fine-grained, yet inherently noisy and prone to spurious correlations. Conversely, knowledge distilled through rule inference systems (e.g., decision trees, fuzzy inference systems, or association rule mining \cite{huang2025PARM}) yields clean, interpretable logic, but often at the cost of losing nuanced, continuous context. The optimal paradigm lies in their synthesis. This interplay between rule-based systems and JEPA offsets the disadvantages of both extremes. Further, by projecting knowledge into shared \textit{embeddings}, rather than attempting to model logic at the raw pixel, observation, or syntactic token level, the architecture inherently filters out low-level stochasticity. This allows the network to focus purely on high-level semantic relationships and stable, causal structures.

\paragraph{Rule Injection as Multi-Level Regularization}
Our empirical observations, particularly the explicit \textit{Logic-Accuracy Trade-off}, highlight the regularizing effect of neuro-symbolic constraints. As demonstrated, a Classic JEPA prioritizes the minimization of continuous data reconstruction error (MSE) by capturing any available statistical correlation. RiJEPA, however, actively trades a fraction of this raw correlational accuracy to strictly satisfy absolute symbolic constraints. 

The theoretical implications of this re-shaping depend on the origin of the rules. If the rules are extracted from the \textit{same dataset} (as in our clinical experiment), RiJEPA acts as a multi-level learning framework. It integrates different granularities of information: fine-grained continuous details and macro-level logical abstractions where the rules serve as a powerful structural regularizer to prevent the manifold from collapsing into entangled, spurious correlations. Conversely, if the rules are derived from an \textit{external source} (such as established medical guidelines or human expert priors), this mechanism serves as a robust vehicle for zero-shot knowledge injection, forcing the model to learn representations that are \textit{coherent} with domain expertise. Our synthetic and real-world experiments confirm that energy-based rule constraints successfully re-shape this latent geometry.

This mechanism explains the dichotomy observed in our linear probing results. While a pure self-supervised JEPA can disentangle continuous features well enough for an \textit{ex-post} downstream supervised classifier to achieve 100\% accuracy, it remains fundamentally blind to the semantic meaning of those features. RiJEPA, guided by structural regularization, intrinsically internalizes the diagnosis, enabling 100\% Zero-Shot logical inference without ever requiring downstream labeled fine-tuning.

\paragraph{Differentiable Logic and Generative Extensions}
The fundamental advantage of this explicitly (re)shaped geometry lies in the realization of \emph{differentiable logic}. By relaxing rigid, discrete symbolic rules into structured, continuous latent embeddings, we bypass the non-differentiable bottlenecks that have historically forced rule discovery into NP-hard combinatorial search. This geometric transformation allows the neural network to compute explicit gradients ($\nabla_z E(z)$) directly through logical constructs. Consequently, as demonstrated by our generative Langevin paradigms, the model can fluidly navigate the rule space via gradient descent, smoothly interpolate decision thresholds, deduce abductive root causes, and algebraically compose complex reasoning.

Because a Classic JEPA lacks this symbolic vocabulary and structural energy boundary, it is mathematically incapable of zero-shot logical inference or generative rule discovery. However, the generative framework established here, specifically the joint context-target distribution (Eq.\ref{eq:joint_context_target_dist}), offers broader implications for representation learning. If a standard multi-modal JEPA (e.g., matching vision and language) were to define an explicit cross-modal energy surface in its shared latent space, \textit{it could potentially adopt our gradient-based Langevin diffusion approach to enable generative discovery across continuous modalities, bypassing the need for traditional autoregressive decoders}.

\paragraph{Future Work: Contrastive Triplet Manifolds}
Scaling this interplay to highly complex, real-world datasets introduces further opportunities for theoretical refinement. Currently, our framework employs an explicit \textit{anchor loss} ($\mathcal{L}_{anchor}$) acting as an absolute centroid alignment: it penalizes the absolute distance between a patient's predicted continuous state and predefined, fixed symbolic poles ($z_{rh}$ for high risk, $z_{rl}$ for low risk). This operates alongside the Energy-Based Constraint ($\mathcal{L}_{EBC}$), which handles positive and negative rule pairs separately. 

Beyond these piecewise constraints, the architecture may be naturally extended to \textit{contrastive triplet training} \cite{schroff2015facenet, weinberger2009distance}. Rather than relying on fixed absolute poles or separated pairs, a triplet loss evaluates the relative distance between three dynamic items simultaneously: an Anchor antecedent ($A$), a valid Positive consequent ($P$), and a corrupted Negative consequent ($N$). This unified loss strictly enforces that the distance $d(A, P)$ remains smaller than $d(A, N)$ by a predefined margin. By training the system directly on these dynamic triplets, we can facilitate \textit{contrastive JEPA rule manifold learning}. This approach would more naturally capture hierarchical rule relationships and fine-grained symbolic compositionality. Future work will explore this contrastive extension, evaluate the practical computational limitations of joint Langevin diffusion in higher-dimensional semantic spaces, and further refine robust latent-to-symbolic rule decoding mechanisms.

\section{Conclusion} \label{sec:conclusion}

In this work, we introduced a comprehensive, bidirectional framework integrating discrete symbolic rules with continuous predictive latent architectures. In Direction 1 (\textit{Rules for JEPA}), we established Rule-Based JEPA (RbJEPA) for pure symbolic distillation and Rule-informed JEPA (RiJEPA) for combining multi-modal knowledge sources. By interpreting rule validity as a geometric energy landscape, RiJEPA injects structured inductive biases to constrain raw-data training, actively mitigating shortcut learning and creating highly robust, out-of-distribution resistant representation spaces. 

In Direction 2 (\textit{JEPA for Rules}), we demonstrated how this explicitly shaped geometry enables generative, continuous rule manifold learning. By transforming rigid, discrete symbols into a continuous, differentiable logic, this framework bypasses the NP-hard bottlenecks of traditional combinatorial search, equipping the network with the ability to organically discover, interpolate, and compose novel rules natively through gradient-guided optimization ($\nabla_z E(z)$).

We empirically validated this dual-direction framework through both controlled topological simulations and a high-stakes clinical case study. Our experiments on the real-world UCI Heart Disease dataset reveal a critical distinction: while classic predictive architectures can learn continuous features that are linearly separable \textit{ex-post}, they can be blind to semantic meaning and rely heavily on downstream supervised classifiers. In contrast, RiJEPA leverages a formal logic-accuracy trade-off to intrinsically align its representation space during pretraining. This explicit topological clustering allows the model to achieve perfect zero-shot logical consistency without downstream fine-tuning, while unlocking the generative flexibility required for complex forward inference and abductive reasoning. Unlike standard black-box models, RiJEPA provides a transparent, distance-based geometric interpretation of its decision-making process, where clinical diagnoses are emergent properties of a logically structured latent manifold. Ultimately, this bridge between raw neural pattern recognition and symbolic expert logic establishes a powerful new foundation for neuro-symbolic representation learning, and offers a highly promising path towards trustworthy, interpretable AI.

\bibliographystyle{plain}
\bibliography{reference}

\appendix

\section{Pairwise Predictive Supervision vs.\ Energy-Based Rule Constraints} \label{app:PPS_vs_EBC}

Here, we formally compare the two supervision strategies discussed in the main text: the simple Pairwise Predictive Supervision (PPS) used for RbJEPA (Eq.~\ref{eq:RbJEPA_loss}) and the explicit Energy-Based Constraint (EBC) utilized in RiJEPA (Eq.\ref{eq:RiJEPA_EBC_loss}).

\subsection{Shared Mathematical Form}

Both implementation strategies rely on the same core geometric quantity:
$$E(A, C) = \left\| g\!\left(f_c(A)\right) - f_t(C) \right\|_2^2$$
where:
\begin{itemize}
    \item $f_c$ is the context encoder,
    \item $f_t$ is the target encoder,
    \item $g$ is the universal predictor,
    \item $A \rightarrow C$ represents a rule with antecedent $A$ and consequent $C$.
\end{itemize}
Mathematically, the expression used to compute the pairwise prediction error is identical to the energy function. The core difference lies not in the formula itself, but in how the optimization landscape is constrained during training.

\subsection{Pairwise Predictive Supervision (PPS)}

In the pairwise strategy (Eq.~\ref{eq:RbJEPA_loss}), the objective optimizes only the positive examples:
$$\min_\theta \sum_{(A, C) \in \mathcal{R}_{\text{valid}}} w_i E(A, C)$$
Only valid rules are considered. There is no explicit modeling or penalization of invalid rule relationships.

\paragraph{Geometric Interpretation \& Implications:}
\begin{itemize}
    \item Valid rule pairs are pulled together in the continuous embedding space.
    \item No explicit repulsive force pushes invalid rules away; the remainder of the latent space is completely unconstrained.
    \item The model learns a standard predictive mapping $g(z_c) \approx z_t$.
    \item It is stable and straightforward to optimize (acting as standard regression).
    \item It does \emph{not} define a bounded rule validity landscape, making it highly susceptible to OOD extrapolation and unsuitable for generative diffusion.
\end{itemize}

\subsection{Energy-Based Rule Constraint (EBC)}

In the energy-based formulation, the same $E(A, C)$ is treated as a learned energy functional over the joint space (Eq.\ref{eq:RiJEPA_EBC_loss}):
$$\min_\theta \sum_{(A, C) \in \mathcal{R}_{\text{valid}}} E(A, C) \quad + \quad \lambda \sum_{(A, C_{neg}) \in \mathcal{R}_{\text{invalid}}} \max\left(0, m - E(A, C_{neg})\right)$$
which equivalently enforces: $E_{\text{valid}}$ must be low, and $E_{\text{invalid}}$ must be high. $m$ is a user-defined separation margin.

\paragraph{Geometric Interpretation \& Implications:}
\begin{itemize}
    \item Valid rules are sculpted into stable low-energy basins, while invalid rules are actively pushed into unstable high-energy regions.
    \item It explicitly models rule validity and defines a strict, separable boundary between valid and invalid logical mappings.
    \item It mathematically constructs a probabilistic generative model over rules: $p(A, C) \propto \exp(-E(A, C))$.
    \item It natively enables gradient-based exploration (e.g., Langevin dynamics) for novel rule discovery.
\end{itemize}

\subsection{Core Conceptual Difference}

Although the scalar function $E(A, C)$ is identical in both formulations, their learning paradigms fundamentally diverge:
\begin{center}
\textbf{Pairwise supervision learns a mapping.} \\
\textbf{Energy-based training learns a landscape.}
\end{center}
Pairwise supervision imposes only \textit{absolute} constraints (pulling valid pairs together). In contrast, Energy-Based training imposes \textit{relative} constraints (ensuring valid mappings have lower energy than invalid ones), thereby actively shaping the global predictive geometry.

\subsection{Summary}

\begin{table}[h]
\centering
\renewcommand{\arraystretch}{1.2}
\begin{tabular}{l|l}
\hline
\textbf{Pairwise Predictive (RbJEPA)} & \textbf{Energy-Based Constraint (RiJEPA)} \\
\hline
Positive-only training & Positive + Negative (contrastive) training \\
Learns a conditional mapping & Learns a rule validity landscape \\
Regression objective & Energy-Based / Margin objective \\
Stable and computationally simple & More powerful but harder to optimize \\
No explicit modeling of invalid logic & Explicit modeling of invalid logic \\
\hline
\end{tabular}
\end{table}

\section{Synthetic Experiment: Simulation Details}
\label{sec:simulation_details}

In this section, we provide the full configuration details for the 3D energy landscape simulation discussed in Section.\ref{subsec:simulated_example} to ensure maximum reproducibility.

\subsection{Data Generation Parameters}
The synthetic environment operates in a 3D feature space ($\mathbb{R}^3$) with the $z$-axis fixed to $0$ for clear 2D visualizations. The dataset generation parameters are defined as follows:
\begin{itemize}
    \item \textbf{Dataset Sizes:} Training observations $N_{data} = 1000$, valid rules $N_{rules} = 200$, negative/invalid rules $N_{neg} = 500$, and testing subsets $N_{test} = 300$ per ID/OOD evaluation.
    \item \textbf{Valid Rule Distributions:} Gaussian centers $\mu_1 = [2.0, 2.0, 0.0]$ and $\mu_2 = [-2.0, -2.0, 0.0]$.
    \item \textbf{Invalid Rule Distributions (OOD):} Gaussian centers $\mu_{neg1} = [-3.0, 3.0, 0.0]$ and $\mu_{neg2} = [3.0, -3.0, 0.0]$.
    \item \textbf{Covariance:} A shared isotropic covariance matrix $\Sigma = 0.5 \mathbf{I}$ was used for sampling rule antecedents. Raw data sampling utilized a scaled covariance of $1.5 \Sigma$ to simulate broader data distributions around the latent rule centers.
    \item \textbf{Logical Transformation:} The ground-truth valid rule mapping was defined as an additive transformation: $C = A + [1.0, 1.0, 1.0]$.
    \item \textbf{Observation Noise:} Isotropic Gaussian noise $\mathcal{N}(0, 0.2^2)$ was added to the raw target data $\mathcal{D}_{raw}$.
\end{itemize}

\subsection{Network Architectures}
The Joint-Embedding Predictive Architecture (JEPA) models consisted of three components: the Context Encoder ($f_c$), the Target Encoder ($f_t$), and the Predictor ($g$). All three were parameterized as identical Multi-Layer Perceptrons (MLPs).
\begin{itemize}
    \item \textbf{Input Dimension:} 3
    \item \textbf{Hidden Dimension:} 32
    \item \textbf{Latent Dimension ($z$):} 16
    \item \textbf{Layer Sequence:} \texttt{Linear(3, 32)} $\rightarrow$ \texttt{GELU} $\rightarrow$ \texttt{Linear(32, 16)}.
\end{itemize}
The predicted energy for a given pair $(x, y)$ was computed as the squared Euclidean distance in the latent space: $E(x,y) = \|g(f_c(x)) - f_t(y)\|^2_2$.

\subsection{Training Hyperparameters}
Both the Classic JEPA and the RiJEPA models were trained using full-batch gradient descent. The pseudo-random number generators for PyTorch \cite{pytorch2019} and NumPy \cite{harris2020array} were seeded ($111$) to ensure reproducibility.
\begin{itemize}
    \item \textbf{Optimizer:} AdamW
    \item \textbf{Learning Rate:} $1 \times 10^{-3}$
    \item \textbf{Training Epochs:} 500
    \item \textbf{EBC Margin ($m$):} 5.0 (Applied exclusively to RiJEPA)
    \item \textbf{EBC Regularization Weight ($\lambda$):} 1.0 (Applied exclusively to RiJEPA)
\end{itemize}

\subsection{Computing Infrastructure}
All experiments, model training, and topological visualizations were executed on a standard cloud-based computing environment (Google Colab \cite{Edwards2024Colab}) with the following hardware specifications:
\begin{itemize}
    \item \textbf{Processor:} Intel(R) Xeon(R) CPU @ 2.20GHz (x86\_64 architecture)
    \item \textbf{Cores:} 2 logical CPUs (1 core per socket, 2 threads per core)
    \item \textbf{System Memory:} 13.61 GB RAM
    \item \textbf{Storage:} 115.66 GB Disk Space
    \item \textbf{Virtualization:} KVM Full Virtualization
\end{itemize}
Given the lightweight nature of the synthetic MLPs and the closed-form dimensionality of the generated data, the entire simulation workflow runs efficiently on the specified CPU infrastructure without requiring dedicated GPU acceleration.

\section{Real-World Experiments: More Details} \label{app:real_world_exp_details}

The implementation of the Multi-Modal Dual-Encoder architecture for the heart disease case study, as in Section.\ref{subsec:clinical_case}, utilizes synchronized latent dimensions to facilitate the Shared Latent Semantic Space.

\subsection{Encoder and Predictor Specifications}
All encoders ($f_{c\_data}, f_{t\_data}, f_{c\_rule}, f_{t\_rule}$) and the shared predictor ($g$) share a standardized bottleneck of $d=32$.

\paragraph{Encoder Architecture ($f$)}
Both data and rule encoders utilize a four-layer MLP structure to project input features into the latent space:
\begin{enumerate}
    \item \textbf{Input Layer:} Dimension $d_{in}$ (variable based on modality).
    \item \textbf{Linear Projection:} $d_{in} \rightarrow 64$.
    \item \textbf{Activation:} Gaussian Error Linear Unit (GELU).
    \item \textbf{Normalization:} LayerNorm(64).
    \item \textbf{Output Projection:} $64 \rightarrow 32$.
\end{enumerate}

\paragraph{Shared Universal Predictor ($g$)}
The predictor is designed to facilitate complex mappings between context and target embeddings. It accepts a latent vector of size 32 and returns a predicted target vector of size 32:
\begin{enumerate}
    \item \textbf{Expansion Layer:} Linear(32 $\rightarrow$ 64).
    \item \textbf{Activation:} GELU.
    \item \textbf{Normalization:} LayerNorm(64).
    \item \textbf{Compression Layer:} Linear(64 $\rightarrow$ 32).
\end{enumerate}

\subsection{Training Hyperparameters}
The framework was trained utilizing the following parameters and regularization strategies:
\begin{itemize}
    \item \textbf{Optimizer:} AdamW optimizer with a learning rate of $1 \times 10^{-3}$.
    \item \textbf{Epochs and Batch Size:} the model was trained for $100$ epochs with a batch size of $32$.
    \item \textbf{Objective Function Weights:} the total loss comprised the JEPA prediction loss, the Energy-Based Constraint (EBC), and an explicit anchor loss aligning vitals with clinical outcomes. The components were weighted as: $\mathcal{L}_{total} = \mathcal{L}_{JEPA} + 2.0\mathcal{L}_{EBC} + 5.0\mathcal{L}_{anchor}$.
    \item \textbf{Target Encoder Momentum:} to prevent representational collapse, the data target encoder ($f_{t\_data}$) was frozen and its weights were updated via an Exponential Moving Average (EMA) from the context encoder, governed by a momentum parameter of $\tau = 0.99$.
    \item \textbf{Gradient Clipping:} gradients were clipped to a maximum norm of $1.0$ (\texttt{clip\_grad\_norm\_}) to ensure optimization stability within the multi-modal loss landscape.
\end{itemize}

\subsection{Shared Latent Bridge Mechanics}
As shown in Fig.\ref{fig:clinical_predictor_architecture}, the predictor $g$ is agnostic to the source modality. Whether the input is a patient embedding from $f_{c\_data}$ or a rule antecedent from $f_{c\_rule}$, the predictor operates on a unified 32-dimensional manifold. This alignment is enforced by the combined loss detailed above. During generative Langevin Discovery, we utilize a fixed noise parameter (temperature) scaled to $0.01$ and a gradient step size of $\eta=0.1$ executed over 100 iterations to efficiently navigate this energy manifold.

\subsection{Baseline Classic JEPA and Fair Comparison}
To ensure a scientifically rigorous and fair baseline evaluation, we implemented a Classic JEPA architecture that is strictly comparable to the proposed RiJEPA framework. The comparison is mathematically sound and carefully controlled based on the following principles:

\paragraph{1. Identical Data Pathway}
The Classic JEPA model acts as an exact clone of the Multi-Modal Dual-Encoder, omitting only the rule-specific components. Both models share an identical parameter count for data processing: they use the exact same Encoder class for $f_{c\_data}$ and $f_{t\_data}$ (a 3-layer MLP ending in a 32-dimensional latent space). Further, they utilize the exact same predictor capacity for $g$ (expanding to 64 dimensions, applying GELU/LayerNorm, and compressing back to 32). Because the capacity of the neural network processing the raw patient data is identical across both models, we entirely eliminate the confounding risk that RiJEPA performs better simply due to a larger data-processing network.

\paragraph{2. Identical Training Data and Masking}
Inside the training loop, both models process the exact same batches of raw patient vitals ($x_t$) and receive the identical corrupted/masked input contexts ($x_c$). Consequently, both architectures compute the exact same baseline data-reconstruction loss: $\mathcal{L}_{JEPA} = \| g(x_c) - x_t \|_2^2$. Additionally, both models update their respective target encoders using the exact same Exponential Moving Average (EMA) momentum ($\tau=0.99$).

\paragraph{3. Isolating the Neuro-Symbolic Variable}
In a rigorous ablation study, it is critical to isolate exactly one variable. Here, the controlled variable is the gradient updates derived from the symbolic rules. While the Classic JEPA receives gradients exclusively from the statistical correlation of the masked data ($\mathcal{L}_{JEPA}$), RiJEPA receives these exact same data gradients \textit{plus} the structural constraints dictated by the rules ($\mathcal{L}_{EBC}$ and $\mathcal{L}_{anchor}$). 

Because the architectures, network capacities, and raw data inputs are completely identical, any improvement in RiJEPA's downstream linear probing accuracy, or any structural clustering observed in its t-SNE manifold, can be strictly and exclusively attributed to the neuro-symbolic rule constraints. This design proves that it is the addition of symbolic logic (rather than arbitrary architectural scaling) that makes the latent space fundamentally more robust and interpretable.

\end{document}